\documentclass{article}
\usepackage{mdframed}
\usepackage{booktabs}
\usepackage{longtable}
\usepackage{titlesec}
\usepackage{enumerate}   
\usepackage{geometry} 
\usepackage[dvipsnames]{xcolor}
\usepackage{caption}
\usepackage{multirow}
\usepackage{multicol}

\titleformat{\paragraph}
{\normalfont\normalsize\itshape}{\theparagraph}{1em}{}
\titlespacing*{\paragraph}
{0pt}{3.25ex plus 1ex minus .2ex}{1.5ex plus .2ex}

\usepackage[utf8]{inputenc} 
\usepackage{tabularx} 
\usepackage{booktabs} 
\usepackage{multirow} %
\usepackage{amsmath}
\usepackage{graphicx}
\usepackage{hyperref}
\usepackage{multirow}
\usepackage[linesnumbered,ruled,vlined]{algorithm2e}
\usepackage{lmodern}
\usepackage{algpseudocode}
\usepackage{geometry}
\usepackage{fancyhdr}
\usepackage{tikz}
\usepackage{booktabs} 

\usepackage{listings}
\usepackage{color}
\usepackage{listings}
\usepackage{xcolor}
\usepackage{lineno}
\usepackage{authblk}

\definecolor{seagreen}{RGB}{46, 139, 87}

\usepackage{amssymb}

\title{Extracting Structured Requirements from Unstructured Building Technical Specifications for Building Information Modeling}

\author[1,3]{Insaf Nahri}
\author[2]{Romain Pinquié}
\author[1]{Philippe Véron}
\author[3]{Nicolas Bus}
\author[3]{Mathieu Thorel}

\affil[1]{LISPEN EA 7515, Arts et Métiers Institute of Technology, 13100 Aix-en-Provence, France}
\affil[2]{Univ. Grenoble Alpes, CNRS, Grenoble INP, G-SCOP, 38000 Grenoble, France}
\affil[3]{Information System and Applications Division, CSTB, 06560 Sophia Antipolis, France}

\begin{document}

\maketitle

\begin{figure}[h!]
\centering
\includegraphics[width=\textwidth]{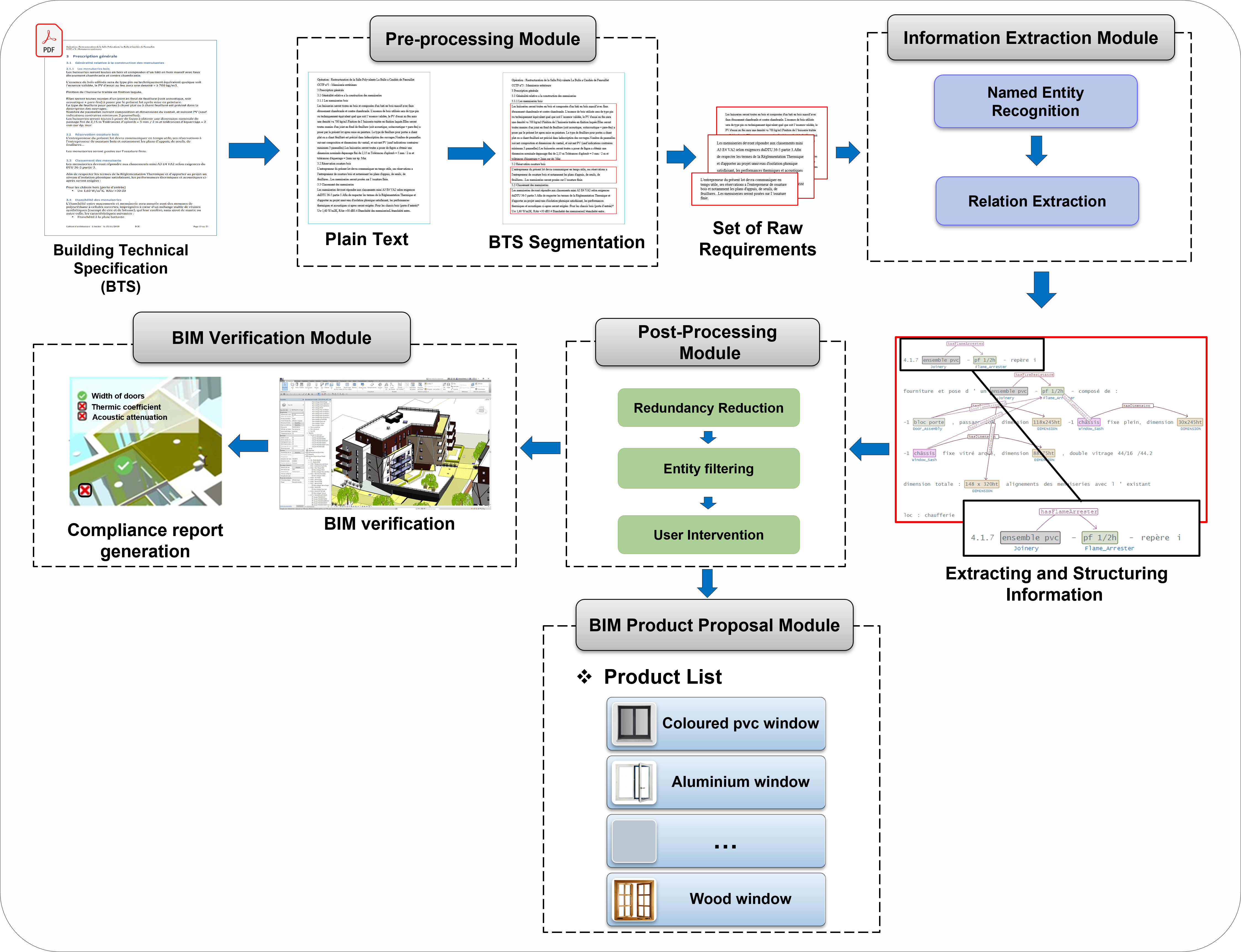}
\caption{Graphical abstract outlining the overall workflow: 1) pre-processing building technical specification documents, 2) extracting raw requirements, 3) identifying named entities and their relationships, 4) defining structured formal requirements, 5) Verifying the BIM digital mock-up and selecting compliant products from catalogues.}
\label{GraphicalAbstract}
\end{figure}

\begin{abstract}
This study explores the integration of Building Information Modeling (BIM) with Natural Language Processing (NLP) to automate the extraction of requirements from unstructured French Building Technical Specification (BTS) documents within the construction industry. Employing Named Entity Recognition (NER) and Relation Extraction (RE) techniques, the study leverages the transformer-based model $CamemBERT$ and applies transfer learning with the French language model $Fr\_core\_news\_lg$, both pre-trained on a large French corpus in the general domain. To benchmark these models, additional approaches ranging from rule-based to deep learning-based methods are developed. For RE, four different supervised models, including Random Forest, are implemented using a custom feature vector. A hand-crafted annotated dataset is used to compare the effectiveness of NER approaches and RE models. Results indicate that $CamemBERT$ and $Fr\_core\_news\_lg$ exhibited superior performance in NER, achieving F1-scores over 90\%, while Random Forest proved most effective in RE, with an F1 score above 80\%. The outcomes are intended to be represented as a knowledge graph in future work to further enhance automatic verification systems.
\end{abstract}



\section{Introduction}

Building Information Modeling (BIM) significantly enhances communication and streamlines workflows in the architectural, engineering, and construction (AEC) sectors through virtual 3D representations of buildings. Despite these benefits, BIM frequently involves the complex management of unstructured documents, such as the French BTS documents, known as \textit{Cahier des Clauses Techniques Particulières (CCTP)}. These BTS documents are pivotal in detailing technical requirements and instructions for BIM model validation. They encompass various work packages necessary for effective execution, where each package is delineated by a BTS document varying in length from 10 to 50 pages, containing hundreds of sentences that describe technical requirements and standards. The number and diversity of BTS work packages widely vary based on project scale, with larger projects featuring multiple BTS work packages addressing different systems such as electrical, joinery, or plumbing, etc. This poses significant information management challenges, forming the primary problem this paper addresses.

Current Automated Code Checking (ACC) systems, crucial for validating BIM models against specified requirements, largely depend on manual information extraction (IE) \cite{zhang2021deep}. Although some ACC systems feature semi-automated processes, they still rely on manual annotations or predefined IE rules, limiting their adaptability and scalability \cite{zhang2016semantic, zhou2017ontology}. In contrast, machine learning-based IE methods utilize models to autonomously discern the underlying syntactic and semantic patterns in the text, offering greater flexibility and scalability than traditional methods. These approaches significantly reduce the initial and ongoing efforts needed to develop and update IE rules. Recent advancements have demonstrated that deep neural networks can effectively learn the complex syntax and semantics characteristic of natural language \cite{liu2019bidirectional,zhang2021deep}, which is particularly beneficial for documents like BTS that often vary in style and structure due to different authors and the complexity of the requirements. Additionally, their specifications are long and hierarchically complex, necessitating sophisticated parsing capabilities. Furthermore, linguistic diversity in construction documents poses a significant challenge, as most existing research focuses on English, with limited attention to other languages. Despite the high performance of machine learning methods in IE, the variability, complexity, and linguistic diversity inherent in construction documents like BTS are not fully addressed by existing systems, which fail to efficiently adapt to the diverse and extensive nature of these documents.

This paper introduces a novel deep learning-based approaches aimed at enhancing ACC systems by automating the extraction of information from French BTS documents. It proposes utilizing Named Entity Recognition (NER) and Relation Extraction (RE) to autonomously detect and analyze syntactic and semantic patterns within these texts. This approach leverages $CamemBERT$, a BERT-based model \cite{martin2019camembert}, and a transfer learning technique using "$Fr\_core\_news\_lg$", both pre-trained on a broad French corpus in the general domain. Also, this study develops various approaches to compare these models, primarily because there are no existing works that extract requirements from French BTS documents. Additionally, this is the first study to explore the use of $CamemBERT$ and "$Fr\_core\_news\_lg$" in the construction domain. Furthermore, this study aims not only to extract building entities but also to identify the relationships between them, thereby creating a comprehensive system for requirement extraction. To achieve this, the study explores four supervised models for RE, such as Random Forest, to determine relationships between entities based on a custom feature set. The results obtained can be represented as a knowledge graph and juxtaposed with the BIM model for model BIM verification.

The paper is organized as follows: Section \ref{stateofart} reviews the literature on IE techniques within the construction industry, emphasizing their applications and limitations. Section \ref{methode} details the methodology, exploring various approaches to NER and RE, tailored to address the complexities of BTS documents. Section \ref{exp} presents a comparative analysis through experimental setups, focusing on the effectiveness of different techniques in domain-specific applications such as the construction industry. Section \ref{conclusion} concludes with a summary of the key findings, discusses the limitations of the approach, and outlines future research directions to enhance ACC systems' capabilities in managing diverse and extensive project documentation in the AEC sector.

\begin{figure}[h!]
\centering
\includegraphics[width=\textwidth]{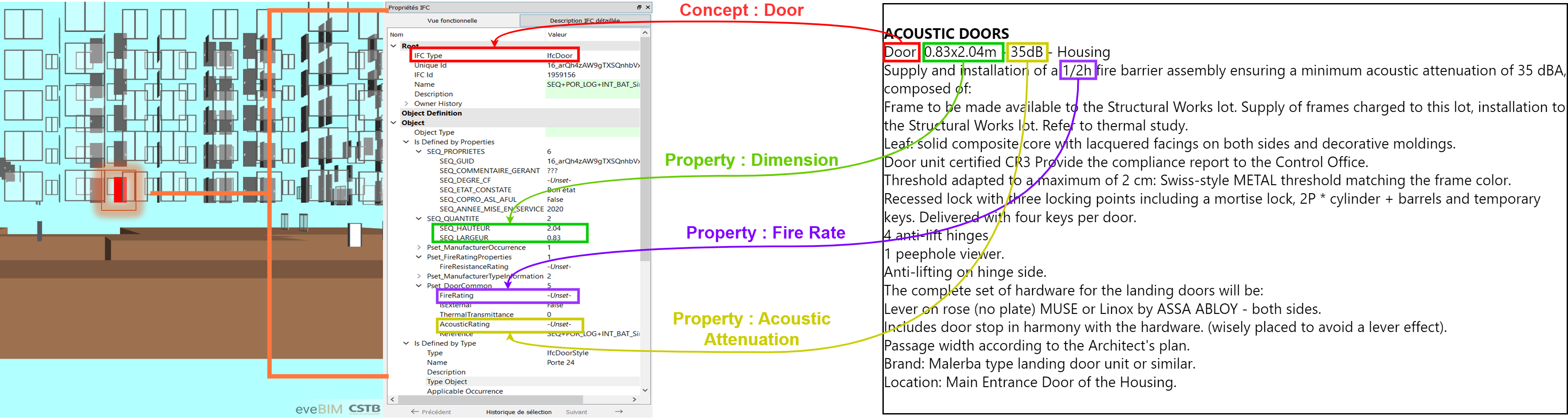}
\caption{The figure on the left shows a BIM model with a list of properties related to the concept of a door, and on the right, a section of the BTS document requirements, translated into English. This section contains the text requirement that the door must comply with, such as the door must have a fire rating of half an hour (1/2h). However, in the BIM model, this value is unset, highlighting the necessity of compliance checking between the BTS requirements and the BIM model.}
\label{BIM&CCTP}
\end{figure}

\section{Literature review}\label{stateofart}
This section reviews NLP techniques for IE within the construction domain (Section \ref{ie_in_construction}) before reviewing various approaches for NER (Section \ref{ner_approach}) and RE (Section \ref{re_approach}), concluding with a discussion (Section \ref{et_discussion}).

\subsection{Information Extraction}\label{ie_in_general}
Information Extraction (IE) is a fundamental component of NLP, dedicated to automating the conversion of unstructured text into structured information across various domains \cite{nismi2023review}. IE facilitates the unlocking of valuable insights within texts, notably through tasks such as NER, which identifies entities like people, organizations, and locations, and RE, which discerns the connections and associations among these entities to underscore significant relationships \cite{jiang2012information}. These processes transform the raw text into analyzable data, crucial for enabling machines to understand and process information. Additionally, IE frequently incorporates ontologies and knowledge graphs, which help structure and semantically enhance the extracted data, further enriching the analysis \cite{ahaggach2023information,fahad2018semantic}.

\subsubsection{Named Entity Recognition for Information Extraction Approaches}\label{ner_approach}
Existing NER methods can be classified into four primary categories: rule-based approaches, dictionary-based approaches, statistical machine learning-based approaches, and deep learning approaches.

\textbf{Rule-based approaches} rely on hand-crafted rules and patterns \cite{hobbs2010information}. These rules are often tailored based on domain-specific knowledge and linguistic patterns, making them suitable for extracting structured information from unstructured text. These patterns can take various forms, such as grammars for parsing text, regular expressions to extract parts of strings, etc. \cite{dalianis2018computational}. While rule-based methods are effective for well-understood problems, they are labor-intensive, require a deep understanding of the problem, and may struggle with unexpected data. This approach has been widely adopted in various domains \cite{datta2019frame,rivas2019automated,pinquie2016natural,deleger2010extracting,rahem2015rule}, including the construction realm \cite{zhang2016semantic,ren2021semantic,wu2022rule,zhou2017ontology}.

\textbf{Dictionary-based approaches} involve the use of predefined dictionaries or lists of terms, phrases, or entities of interest. These dictionaries contain specific words that the IE system seeks to identify and extract from the text. Dictionary-based approaches are particularly useful for extracting known entities, such as names of people, locations, or product names, from unstructured text. Dictionary-based approaches were widely used because of their simplicity and their performance in various domains, exemplified by studies in \cite{qiu2020dictionary,quimbaya2016named,fu2023biomedical}. In the context of the construction field, pertinent research can be observed in \cite{akanbi2021design,fan2015project}. 

\textbf{Statistical machine learning-based approaches} to IE utilize statistical models and algorithms to automate the classification and extraction of structured information from unstructured text, primarily through NER. In these methods, NER is viewed as a type of text classification \cite{freitag2000machine} and often approached as a sequence labeling problem where the goal is to determine the optimal label sequence for each input sentence \cite{lei2014comprehensive}. These approaches leverage machine learning models to detect patterns, relationships, and regularities in labeled training data. They employ statistical techniques to predict the presence and structure of data within the text, making them particularly effective at managing unstructured data sets. Common techniques used include Hidden Markov Models \cite{morwal2012named}, Conditional Random Fields (CRFs) \cite{skeppstedt2014automatic,patil2020named}, Maximum Entropy Markov Models \cite{alam2020proposed}, and Support Vector Machines (SVMs) \cite{hamad2023medical}. However, these methods often require substantial labeled datasets and careful feature engineering, which can be resource-intensive and limit their scalability and adaptability to new domains \cite{wang2019incorporating}. 

\textbf{Deep learning-based approaches} has significantly advanced IE and NER, eliminating the need for hand-crafted features. A key example in the NER domain is the BiLSTM-CRF model, which leverages the strengths of Bidirectional Long Short-Term Memory (BiLSTM) networks combined with CRF to achieve precise and structured sequence labeling. This architecture captures intricate contextual information through the BiLSTM layers and refines predictions with the CRF layer by considering label dependencies and ensuring consistency, making BiLSTM-CRF highly suitable for complex tasks like NER \cite{gridach2017character,ji2020power}. One of the principal challenges in applying deep learning to IE is the substantial need for extensive annotated text datasets. Standard resources such as the Penn Treebank datasets for syntactic and semantic analysis \cite{marcus1993building}, CoNLL-2003 for language-independent NER \cite{sang2003introduction}, and the $SUD\_French-Sequoia treebank$, which offers a deep syntactic representation scheme for French \cite{candito2014deep}, are typically oriented towards general NLP tasks rather than specific IE applications. The scarcity of domain-specific annotated data, particularly in sectors like AEC, poses significant hurdles for deploying deep learning effectively in these areas.

In response to these challenges, the field has embraced transfer learning and Large Language Models (LLMs) based on transformer architectures. Transfer learning improves domain-specific IE by reducing the effort and cost associated with preparing annotated training data \cite{tan2018survey}. For instance, the $SUD\_French-Sequoia treebank$ has been instrumental in training models like spaCy's “$Fr\_core\_news\_lg$”, which utilizes a Convolutional Neural Network (CNN) architecture designed for efficient and fast processing of French text, capturing syntactic and semantic nuances through word features and pre-trained word vectors for tasks like POS tagging, NER, and dependency parsing \cite{spacy_101}.

The landscape of language models has been transformed by the introduction of transformer-based Masked Language Models (MLMs) such as Google's bidirectional encoder representations from transformers ($BERT$) \cite{devlin2018bert}, trained on an English corpus, and its French derivatives, $CamemBERT$ \cite{martin2019camembert} and $FlauBERT$ \cite{le2019flaubert}. $CamemBERT$, introduced by Martin et al. in 2018, and $FlauBERT$, introduced by Hang et al. in 2019, are both BERT-derived models tailored to the French language, trained on extensive French corpora. These models adapt BERT's powerful transformer architecture, enhancing their ability to capture contextual subtleties for precise domain-specific entity extraction. Their effectiveness is further augmented by attention mechanisms \cite{ashish2017attention}, which streamline the training process compared to traditional Recurrent Neural Networks (RNNs) and allow for nuanced understanding of text without extensive labeled datasets \cite{zheng2022pretrained,devlin2018bert}. Moreover, Causal Language Models (CLMs) like OpenAI's generative pre-trained transformer (GPT) \cite{radford2018improving} have introduced "In-context learning" (ICL) methodologies \cite{wies2024learnability} that support zero and few-shot learning scenarios, where zero-shot learning occurs without any specific examples, and few-shot learning with only a few. Unlike MLMs that predict masked words, CLMs generate text based on prompts, with minimal examples. This innovative approach is a current research focus due to its potential in improving NER performance \cite{zhang20232iner}. The ongoing comparative study of MLMs and CLMs in specialized NER applications highlights the rapid evolution of this technology and its expanding applicability across diverse sectors.

\subsubsection{Relation Extraction for Information Extraction Approaches}\label{re_approach}
Relation Extraction (RE) is a crucial task within IE that identifies and categorizes relationships between entities, which in this context, are specific BIM concepts \(\mathcal{C}\) and properties \(\mathcal{P}\). Relationships are defined by a predefined set of types \(\mathcal{R}\), predicting connections such as “$hasThermicCoe\!f\!ficient$”, “$hasDimension$” among others. RE methodologies are primarily divided into rule-based and machine learning-based categories, with the latter further subdivided into supervised, semi-supervised, distantly-supervised, and unsupervised approaches.

\textbf{Rule-based RE} depends on predefined linguistic patterns and heuristics to derive relationships from texts, using structures like dependency parse trees to extract relations, for example, between proteins \cite{fundel2007relex}. While effective within specific domains, these methods require extensive domain knowledge and struggle with scalability and adaptability.

\textbf{Supervised RE} require a substantial amount of training data to train a classifier that categorizes entity pairs into predefined relation types. There are two types of supervised methods \cite{bach2007review}: 
\begin{itemize}
    \item Feature-based methods, which utilize both syntactic features (e.g., Part-of-speech (POS) tagging) and semantic features (e.g., the path between the two entities in the dependency tree) to form a feature vector for classification \cite{kambhatla2004combining,rink2010utd}.
    \item Kernel-based methods that use string kernels to measure the similarity between two entities by examining the count of shared subsequences \cite{zhang2008exploring}.
\end{itemize}

\textbf{Semi-supervised RE} applies bootstrapping methods to expand on seed instances in contexts where labeled data are limited, leveraging unlabeled data effectively \cite{etzioni2014open}. 

\textbf{Distantly-supervised RE} uses knowledge bases to automatically label text data, assuming that text mentioning known related entities likely describes their relationship \cite{mintz2009distant,quirk2016distant,qin2018robust}.\textbf{Unsupervised RE} identifies and categorizes relationships without reliance on labeled data, using methods like clustering and the K-means algorithm to infer semantic links \cite{hasegawa2004discovering,chen2005unsupervised}. 

\textbf{Deep learning for RE} span various learning paradigms such as supervised, semi-supervised, weakly supervised, and unsupervised, adapted based on implementation and data characteristics. These models leverage neural architectures like CNNs \cite{kumar2017survey,li2018multi}, RNNs \cite{zhang2015relation}, Long Short-Term Memory (LSTM) networks  \cite{xu2015classifying}, and BiLSTM networks \cite{zhang2015bidirectional,zhan2020span} to extract complex patterns and semantic relationships from text without relying on hand-crafted features. Advanced methods such as transformers \cite{nayak2020effective,han2021transformer,ro2020multi} enhance capability further by capturing both local and global textual dependencies. The integration of transfer learning and domain-specific adaptations alongside innovations like few-shot learning in specialized domains, such as medicine \cite{fabregat2023negation}, underscore the significant advances and versatility of deep learning in RE.

\subsection{Information extraction in the construction industry}\label{ie_in_construction}
The construction industry employs IE tasks in various processes, notably ACC. This involves extracting rules from regulatory texts, design standards, and instruction manuals and formalizing them into machine-readable formats. The emergence of NLP techniques has significantly advanced IE. One approach uses a semantic, rule-based NLP method for automated extraction from construction regulatory documents \cite{zhang2016semantic}. Xu et Cai. \cite{xu2019semantic} propose a semantic frame-based information extraction method to parse rule information from the Indian utility accommodation policy. Additionally, an approach for automated building code compliance checking validates Industry Foundation Classes (IFC) model inputs with building code concepts \cite{wu2022model}. Guo et al. \cite{guo2021semantic} suggest a comprehensive semantic ACC process, utilizing NLP to extract rule terms and logical relationships from regulatory documents. Wu et al. \cite{wu2022model2} develop an AEC object identification algorithm using invariant signatures, essential for automated building design model validation for code compliance. Wu et al. \cite{wu2022rule} describe a rule-based method to automatically extract information from mechanical, electrical, and plumbing documents, employing a suffix-based matching algorithm for NER and a dependency-path-based matching algorithm on dependency trees for relationship extraction.

In parallel, machine learning algorithms have become prominent in the field. Zhang and El-Gohary \cite{zhang2021deep} introduce a deep neural network-based method for IE from AEC regulatory documents using BiLSTM-CRF and transfer learning to extract entities and their relationships. Schönfelder and König \cite{schonfelder2021deep} employ a supervised deep learning transformer model (BERT) to extract pertinent terms from a collection of regulatory documents in German. Moon et al. \cite{moon2021automated} automate the review of construction specifications using NLP, developing a NER model with a BiLSTM architecture, and recognizing bridge damage in inspection reports using a recurrent neural network trained with active learning \cite{moon2020bridge}. Another approach involves analyzing semantic properties using NLP techniques, incorporating an NER model based on BiLSTM with a CRF layer \cite{moon2022automated}.

\subsection{Discussion}\label{et_discussion}
IE in the construction industry has transitioned from traditional rule-based methods to advanced deep learning models. While rule-based methods show promising results, they suffer from scalability issues and require significant time to develop rules. Deep learning models, on the other hand, excel at capturing syntactic and semantic features automatically, leading to superior outcomes. However, these models demand large volumes of data, which incurs substantial costs and time to collect, particularly since annotated corpora in the AEC domain are scarce, unlike in the healthcare domain where some annotated corpora for IE are available. This discrepancy has driven recent research towards transfer learning, such as demonstrated by Zhang and El-Gohary \cite{zhang2021deep}, and transformer techniques using BERT, as shown by Schönfelder and König \cite{schonfelder2021deep}, to address these limitations and achieve effective results.

However, these methods face challenges in certain contexts due to linguistic diversity and varying types of regulatory documents across countries. For instance, models like the BiLSTM-CRF in \cite{zhang2021deep} were originally trained on general English data and fine-tuned on domain-specific English regulatory documents. Similarly, transformer-based models like BERT, as used in the study by Schönfelder and König \cite{schonfelder2021deep}, were initially trained on a vast general domain German corpus (160GB) and subsequently fine-tuned on specific German regulatory texts. In contrast, for the French domain, a BERT model named $CamemBERT$, trained from scratch on a 138GB French general domain corpus, exists, but no existing work has leveraged this model for automating IE from French regulatory documents in the AEC domain, such as the BTS documents, to support ACC. Additionally, the Spacy team offers a comprehensive French model, "$Fr\_core\_news\_lg$", trained on a broad French corpus in the general domain, yet no studies have explored this model through transfer learning for domain-specific adaptation in French construction.


Furthermore, while existing deep learning methods process requirements typically formatted as single sentences \cite{zhang2021deep,schonfelder2021deep}, BTS documents often scatter related entities such as concepts and properties across the text. This dispersion requires the detection of text fragments beyond simple sentence boundaries to accurately identify the limits of each requirement and the relationships between different entities. Also in the context of RE, Zhang and El-Gohary try to extract relationships between entities for complete extraction, but in the regulatory text in which they work with the relationship is explicitly written within the text and they handle it as an entity e.g., \textit{“Door openings between a private garage”} \cite{zhang2021deep}; the relationship “between” is explicitly written and can be handled by considering it as an entity. In the BTS case, the relationship is not always explicit but can be included based on the semantic understanding of the text and the type of the properties, especially since the text requirement may be too long and so the distance between entities may also be too long e.g., \textit{"Door assembly of type T or similar. Requested characteristics: - acoustic attenuation R = 53 dB(A) justified by test report"}. Here the relationship between “door assembly” and “R=53 dB(A)” is not evident, necessitating an understanding of the semantic of the text and also to understand the type of the properties which help to capture the implicit relationships. 

To address these needs, This study explores new French NER methods to support ACC in French construction by harnessing LLM-based models illustrated by $CamemBERT$ and the transfer learning of the deep learning model "$Fr\_core\_news\_lg$", both fine-tuned on a labeled BTS dataset to adapt them to the French AEC domain. it compares these models with various NER approaches, ranging from rule-based to deep learning-based, developed for this study, as there is no existing work that addresses BTS documents, distinguishing this research from others. Additionally, while rule-based NER typically uses manual or semi-automated evaluations \cite{guo2021semantic,ren2021semantic, wu2022rule}, this study introduces a method for automating the validation of all proposed approaches. Moreover, this study aims to extract not only building entities but also relationships by addressing the challenges presented by BTS documents. It develops four different models, such as Random Forest, to extract relationships between entities based on custom features. The output of the IE system will be presented as semi-formal requirements, which will be readily formalizable in a knowledge graph format. This format is designed to be compliant with data graphs generated from BIM models and BIM ontologies, such as IfcOWL (Web Ontology Language representation of the IFC) \cite{wang2024enhancement}.

\section{Method}\label{methode}
\begin{figure}[htb!]
\centering
\includegraphics[width=\textwidth]{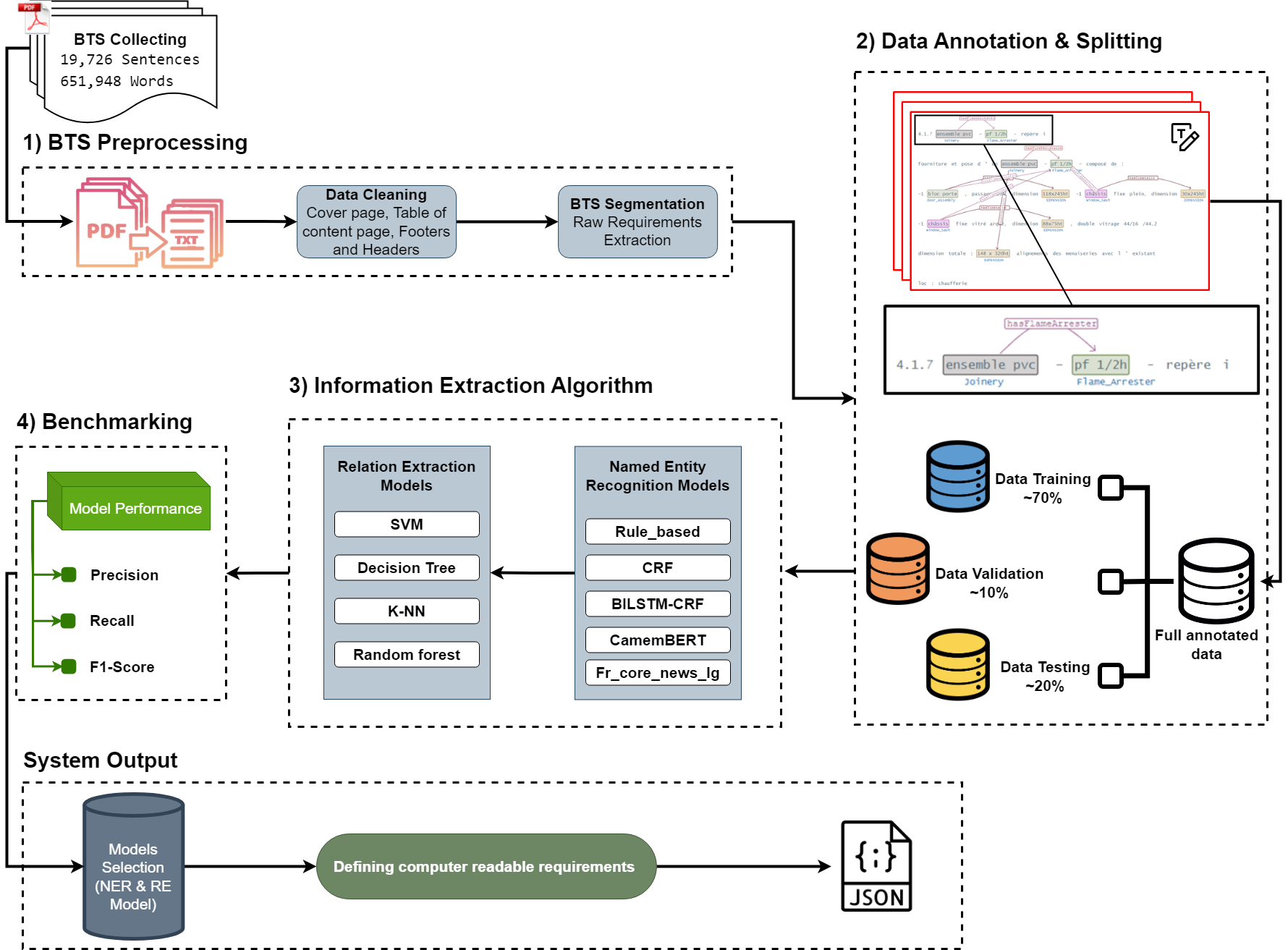}
\caption{Overview of the proposed IE system to extract structured requirements from unstructured BTS.}
\label{DetailedWorkFlow}
\end{figure}

The workflow illustrated in Figure \ref{DetailedWorkFlow} outlines the methodology employed to extract computer readable requirements from unstructured BTS documents. The initial phase involves collecting a set of BTS documents. Following this, we filter out non-relevant information, acting as noise for our information extraction algorithm, during the preprocessing step. Subsequently, we apply a segmentation algorithm to extract raw requirements, encompassing various entities such as concepts and properties that together form a formalized requirement.
In the next stage, we deploy the proposed NER and RE approaches to extract BIM entities and their interrelationships. The outcomes derived from NER and RE are subsequently structured into a JSON file, which can be presented as a knowledge graph to support ACC for further work.

\begin{figure}[h!]
\centering
\includegraphics[width=\textwidth]{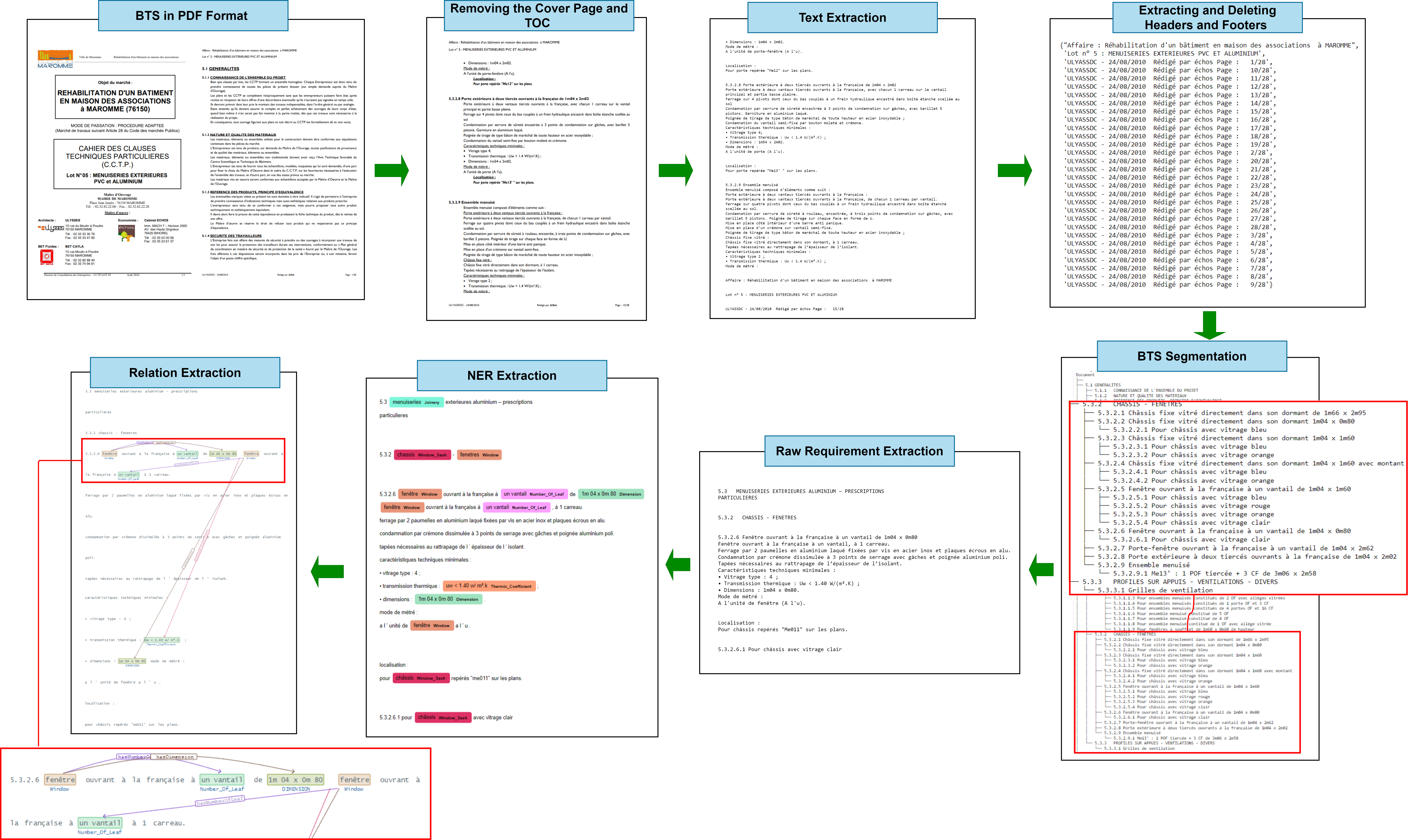}
\caption{Example results of requirement extraction processes on a randomly selected document in the BTS corpus.}
\label{DetailedWorkFlow2}
\end{figure}

Figure \ref{DetailedWorkFlow2} illustrates the output at each step of our requirement extraction system on a randomly selected document within the BTS Corpus. The $Fr\_core\_news\_lg$ NER model is employed for NER and RE, with the utilization of the random forest algorithm.

\subsection{BTS collecting} 
An expert agent conducted the BTS collection process to ensure that the collected BTS documents are representative samples, covering different project types. The BTS documents were in XML-based PDF format (not scanned), comprising a total of 1,505 pages and 651,948 words across 19,726 sentences. They were authored by 61 different individuals, which aids in creating a model capable of extracting formal requirements regardless of the author's style. This is crucial in public contracts, where BTS documents are typically authored by various contractors and building owners, leading to diverse writing styles.

The analysis by the expert agent revealed that a BTS specifies numerous requirements using varied terminology and a broad array of entities, ranging from generic terms such as 'Door' to more specific ones like 'Handle', 'Strap hinge', and 'Frame'. Moreover, the agent revealed that the raw requirements might span multiple sections and subsections. For example, the entities required to construct comprehensive formal requirements could be scattered across several sections of a BTS, not merely at the sentence level. After reviewing the BTS set, it was discovered that the entities are organized hierarchically. Details of this hierarchical organization will be provided in Section \ref{btsseg}.

\subsection{Pre-Processing}
After extracting text from the PDF using specific Python libraries such as "$\text{MuPDF}$\footnote{MuPDF: \url{https://pymupdf.readthedocs.io/en/latest/recipes-text.html}}" and "$\text{pdftotext}$\footnote{pdftotext: \url{https://pypi.org/project/pdftotext/}}", which provide good accuracy in preserving the original formatting and layout of the text within the PDF, the preprocessing step filters out non-relevant information in a BTS. This includes cover pages, the table of contents (TOC), blank pages, footers, and headers.

Deleting the first page of each BTS has resulted in the removal of cover pages, given that the cover page is always the first page of a BTS. The Table of Contents was removed based on the results provided on the extraction of hierarchical structure of BTS discussed in Section \ref{btsseg}.

In both BTS documents and other PDF documents, headers and footers are commonly present. Headers typically appear at the top of pages and include information like document titles, or dates, which are often consistent across documents. Footers are usually located at the bottom and contain elements such as page numbers. These elements can pose challenges in text extraction as they may be intermixed with the main content if not properly segmented, complicating tasks like automated data extraction if they are not appropriately handled. Headers are typically placed at the top of a page, within the top margin, while footers are positioned at the bottom, within the bottom margin. These placements help to separate them from the main body of the text, ensuring they are distinct. Headers and footers are generally consistent throughout a document; however, there are exceptions. For instance, title pages, blank pages, and special pages (such as those with large figures) might omit them. Considering these factors, this paper presents an algorithm designed to automate the extraction and suppression of headers and footers in PDF documents.

Algorithm \ref{alg:identify-repeated-sentences} presents the pseudocode for removing headers and footers from PDF documents. This algorithm successfully extracts headers and footers from 92\% of the documents collected. The challenge arises in the remaining 8\%, primarily due to inconsistencies in headers or footers across different pages. Often, these inconsistencies occur when contractors or building owners inadvertently omit these elements from some pages.

\begin{enumerate}[(i)]
    \item \textbf{Assigning Line Indexes}: The algorithm starts by organizing lines from the first page into two lists: one in normal order (forward) and the other in reverse (backward). This method helps capture headers at the top and footers at the bottom. This step is crucial due to varying line lengths across pages, affecting footer placement. An index “$i$”, ranging from 0 to the length of sentences - 1, assigned to each sentence on each page.
    
    The reason behind assigning this index is to facilitate the comparison of sentences. It allows us to compare, for instance, the sentence “$Si$” on the first page (page 0) with the sentence “$Si$” on every subsequent page “$k$”, where “$k$” belongs to the range \([1, \ldots, n-1]\), and “$n$” represents the total number of pages in the PDF document. This approach ensures the comparison of sentences occupying the same position on different pages, since headers and footers share consistent coordinates and similarity across all pages.
    \item \textbf{Levenshtein Distance for Consistency}: The algorithm utilizes the Levenshtein distance \cite{Cuelogic2023Levenshtein} to compare sentences across pages, aiming to identify consistent ones. A threshold of 5 is chosen to accommodate variations like page numbers. This ensures that sentences in the same position on different pages remain similar, indicative of headers or footers. Additionally, if a figure is detected on a page and the Levenshtein distance is more than 5, indicating a special page with a figure, the comparison is skipped for that page while analysis continues for subsequent ones to maintain header/footer consistency.
    We can express the $Levenshtein$ distance requirement mathematically as follows: For each page, denoted as “$k$”, where $k$ ranges from 1 to $n-1$ (n being the total number of pages), and for each line, denoted as “$i$”, where $i$ represents the line index on each page $k$, ranging from 0 to $m-1$ (m being the total number of lines on page $k$), the $Levenshtein$ distance between sentence $Si_0$ (on page 0) and sentence $Si_k$ (on page $k$) should be less than or equal to 5:
    \[ \forall k \in \{1, 2, ..., n-1\}, \forall i \in \{0, 2, ..., m-1\}: \textit{Levenshtein}(Si_0, Si_k) \leq 5 \]
    In simpler terms, this expression asserts that, for every pair of corresponding sentences between the first page and any subsequent page, the $Levenshtein$ distance should be limited to a maximum value of 5.
    \item \textbf{Storing Repeated Sentences}: Sentences meeting the repetition criteria are stored for further processing and eventual removal.
\end{enumerate}

\begin{algorithm}
\caption{Identify Headers and Footers with Figure Verification}
\label{alg:identify-repeated-sentences}

\KwData{pdfPath: Path to the PDF file}
\KwResult{HeadersFootersList: Set of headers and footers}
  \BlankLine
  \tcp{Try to open the PDF document}
  doc $\gets$ open(pdfPath)\;
  pageTextList $\gets$ Extract text from each page in doc\;
  figurePresenceList $\gets$ CheckForFiguresInEachPage(doc)\; \tcp{Step to check for figures in pages}

  HeadersFootersList $\gets$ set()\;

  \tcp{Concatenate lines of the first page in both normal and reversed order}
  HeadersFootersLinesRef $\gets$ Concatenate lines of the first page in both normal and reversed order\;

  \ForEach{i, sentence \textbf{in enumerate}(HeadersFootersLinesRef)}{
    \If{The stripped sentence $\neq$ “"}{
      isRepeated $\gets$ \textbf{True}\;

      \ForEach{j \textbf{in range}(1, \textbf{len}(pageTextList))}{
        pageText $\gets$ Extract text from the page $j$\;

        \tcp{Concatenate lines of the other pages in both normal and reversed order}
        pageTextLines $\gets$ Concatenate lines of the page $j$ in both normal and reversed order\;

        \tcp{Compare sentence with the corresponding line in the same position on the page using $Levenshtein$ distance}
        \If{$Levenshtein$.distance(sentence, pageTextLines[i]) $\leq$ 5}{
          \If{The stripped pageTextLines[i] $\neq$ “"}{
            \textbf{continue}\;
          }
        }
        \Else{
          \tcp{Check if the current page has a figure}
          \If{figurePresenceList[j]}{
            \textbf{continue}\; \tcp{Continue if there is a figure, as it might disrupt header/footer consistency}
          }
          \Else{
            isRepeated $\gets$ \textbf{False}\;
            \textbf{break}\;
          }
        }
      }

      \If{isRepeated}{
        \ForEach{pageText \textbf{in} pageTextList}{
          Add the stripped line at position $i$ of pageText to HeadersFootersList\;
        }
      }
    }
  }

  \Return HeadersFootersList\;

\end{algorithm}

\subsection{BTS Segmentation}\label{btsseg}
A formal requirement is defined as \(\mathcal{R}q = (\mathcal{C}, \mathcal{R}, \mathcal{P})\), where:
\begin{itemize}
    \item \(\mathcal{C}\) represents concepts, referring to fundamental building elements such as walls, doors, windows, and other structural components in BIM. These concepts provide the backbone for detailed project schematics and planning.
    \item \(\mathcal{R}\) delineates the relations between these concepts and their properties, specifying how each building element (e.g., a door) interacts with its attributes in a manner that aligns with design and construction standards.
    \item \(\mathcal{P}\) comprises properties describing the specific characteristics or attributes of the building elements, such as dimensions, material specifications, and thermal resistance. Each property is represented as a triplet containing a property name, an operator, and a property value, tailored to define the precise requirements necessary for accurate modeling and adherence to construction norms.
\end{itemize}

A BTS contains numerous requirements. The BTS requirements follow a hierarchical pattern, where shared properties are found in the first paragraph. In contrast, specific properties are detailed in subsequent sub-paragraphs. This hierarchy is observed with shared properties typically outlined in the first paragraph, whereas more specific properties are detailed in subsequent sub-paragraphs, reflecting a structured and systematic approach to documentation and analysis.

\begin{figure}[h!]
\centering
\includegraphics[width=\textwidth]{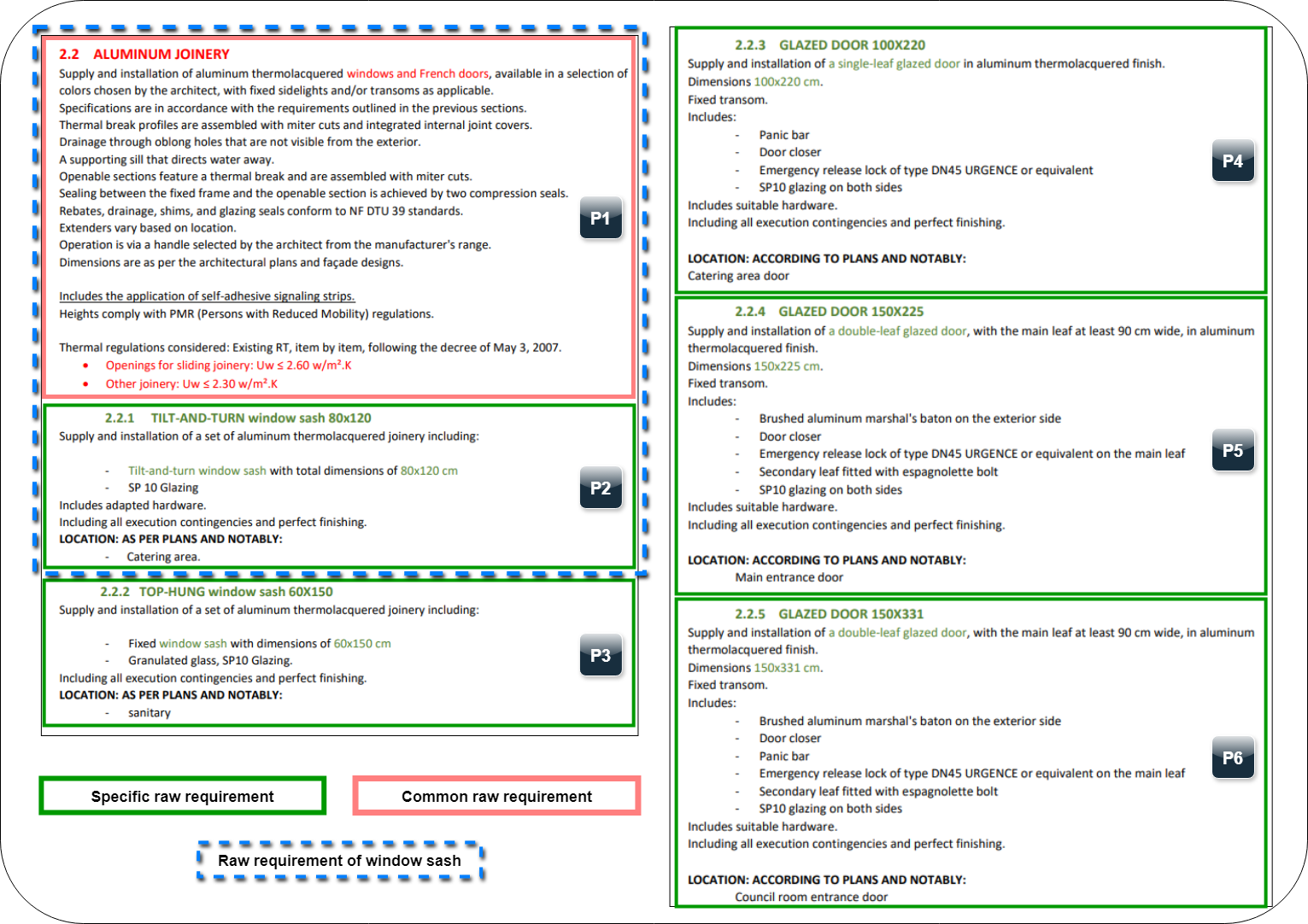}
\caption{Excerpt from the BTS document translated to English, illustrating the role of the BTS hierarchy in extracting complete raw requirements.}
\label{ILPAR}
\end{figure}

Figure \ref{ILPAR} provides an illustrative example using a chapter about aluminum joinery. This chapter is divided into six paragraphs (\(P1, P2, P3, P4, P5,\) and \(P6\)). The distinction among these paragraphs lies in the content they hold. \(P1\) includes specifications that apply to \(P2, P3, P4, P5,\) and \(P6\), making it a “Common Raw Requirement”,For example, it specifies the value of the thermal coefficient that must apply to openings for sliding joinery and other types of joinery, as highlighted in red in the last two lines of \(P1\). \(P2\) and \(P3\) focus on two different window sashes, while \(P4, P5,\) and \(P6\) refer to three distinct door specifications, making them “Specific Raw Requirements” for their respective products. Each specific raw requirement also contains specifications, some of which are highlighted in green in the figure. However, the whole entities that combine all the specifications of a requirement can only be done if the paragraphs that present the common and specific raw requirements are combined to ensure that all the entities required to formalize a requirement are present in the same raw text. This illustrates the necessity of segmenting the BTS based on the hierarchical structure of the BTS.

\begin{figure}[h!]
\centering
\includegraphics[width=\textwidth]{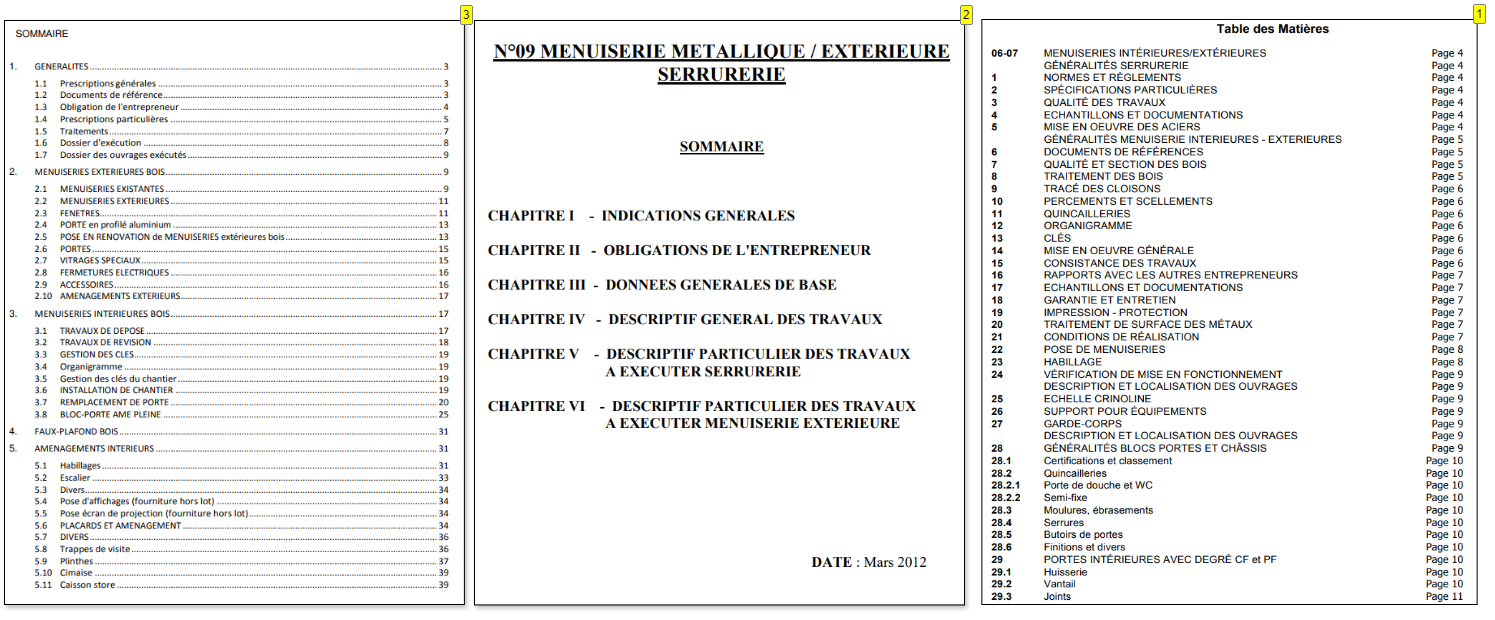}
\caption{TOC randomly selected from our BTS corpora.}
\label{Toc}
\end{figure}

Extracting the hierarchical structure present in PDF documents is necessary to extract complete formalized requirements, that is, combining all concepts, properties, and relationships found in \(P1\) with other paragraphs (\(P2, P3, P4, P5\), and \(P6\)). An initial manual dataset analysis revealed that each BTS does not consistently use font styles, sizes, or colors to differentiate between simple text paragraphs and headlines, which often indicate the start of a new section or subsection. Alternative methods was explored to extract the hierarchical structure of the document through TOC, such as attempting to convert PDFs to Microsoft Word and HTML formats to detect TOC tags to extract the TOC, utilizing packages such as “$docx$\footnote{python-docx: — python-docx 1.1.0: \url{https://python-docx.readthedocs.io/en/latest/}}”, “$pdf2docx$\footnote{pdf2docx: \url{https://github.com/dothinking/pdf2docx}}”, “$aspose-words-cloud$\footnote{Aspose.Words Cloud: \url{https://docs.aspose.cloud/words/}}”, and “$pywin32$\footnote{pywin32: Python for Window Extensions: \url{https://github.com/mhammond/pywin32}}” which provides access to the Windows APIs from Python such as Microsoft Word. However, these approaches proved ineffective because the PDF documents in our dataset might have been created with tools other than Microsoft Word. The transformation from PDF to HTML led to difficulties extracting headlines, which, in turn, hindered the detection of TOC tags. Furthermore, 41.66\% of BTS documents don't include TOC, The presentation of TOC lacks a consistent pattern that would allow for automated removal, as illustrated in Figure \ref{Toc}.

However, a review of BTS showed that principal contractors or building owners frequently use numbering systems to denote document hierarchy, such as \(1, 1.1, 1.1.1, 2,\) or Roman numerals like \(I, II, III,\) or combinations thereof. Therefore, we tried constructing a regular expression to identify the most commonly used numbering systems. The regular expression extracts:
\begin{itemize}
    \item Chapter Heading: Identifies the beginning of a chapter in the text.
    \item Chapter Number: Represents the numerical value assigned to a chapter.
    \item Paragraph Heading: Marks the start of a new paragraph within the text.
    \item Paragraph Number: Indicates the numerical designation of a paragraph.
    \item Section Heading: Signifies the commencement of a new section in the content.
    \item Section Number: Displays the numerical identifier for a section.
    \item Article Heading: Marks the initiation of a new article within the text.
    \item Article Number: Depicts the numerical value assigned to an article.
    \item Enumeration: Represents a sequential list, often denoted by Roman or Arabic numerals.
    \item Sub-enumeration: Indicates a secondary level of enumeration within a list.
\end{itemize}

This approach allowed us to automatically recognize headlines and treat the text between headlines as ordinary paragraphs. We analyzed the numbers associated with each headline to determine its level. For instance, encountering \(1\) or \(I\) would indicate Level 1, and subsequent headlines like \(1.1\) or \(I.1\) would signify Level 2, and so on.

\begin{figure}[hbt]
\centering
\includegraphics[width=0.8\textwidth]{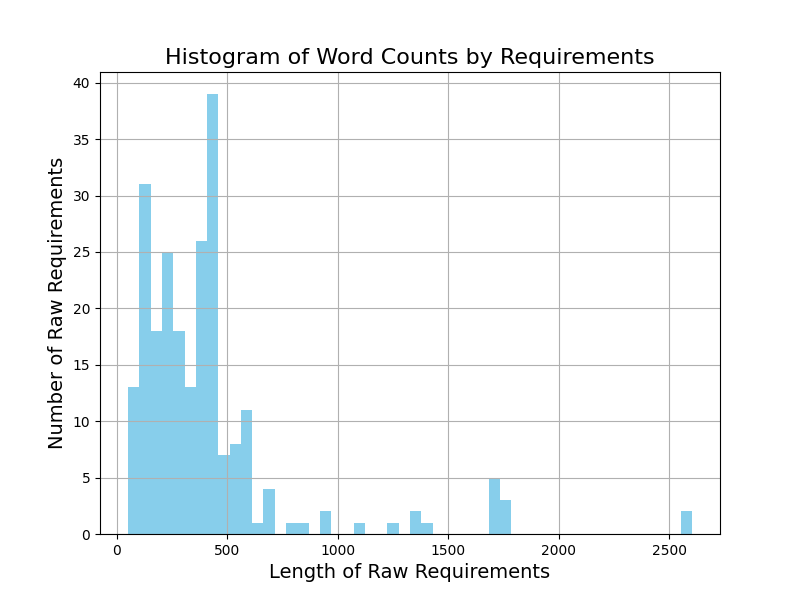}
\caption{Histogram of word count distribution by requirement.}
\label{wc}
\end{figure}

The output of the BTS segmentation algorithm is a list of chunks of text, or combinations of paragraphs, containing the essential components (\(\mathcal{C}, \mathcal{P}, \mathcal{R}\)) necessary to get formal requirements. We will refer to these text blocks as “Raw requirements”. Figure \ref{wc} shows the word count distribution per requirement. The three longest requirements contain a total of 2603 words.




\begin{figure}[ht!]
\centering
\includegraphics[width=0.8\textwidth]{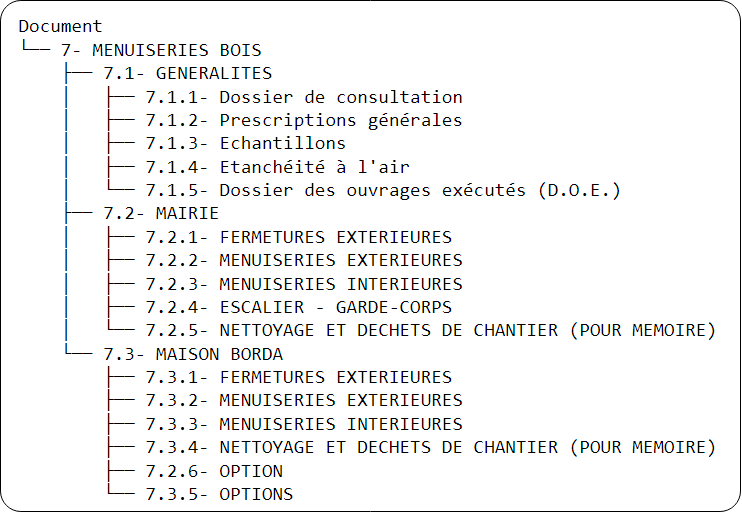}
\caption{Simplified result of extracting Raw Requirements based on the document's hierarchical structure.}
\label{srerr}
\end{figure}

The BTS segmentation algorithm operates in a two-step process. First, it extracts each headline, along with its associated level and paragraph. Then, it merges paragraphs from sibling nodes into their parent nodes. Algorithm \ref{alg2} illustrates the pseudocode for extracting raw requirements from a BTS. Table \ref{TableSegmAnalysis} summarizes how the algorithm performed on various aspects of the corpus, detailing both the successes and challenges faced due to document formatting issues. Figure \ref{srerr} provides a simplified view of the resulting output. As shown in the figure, extracting the hierarchical structure not only enables segmentation of the BTS but also facilitates the regeneration of the TOC of a PDF, which aids in matching and subsequently removing it from the text.

\begin{table}[h]\label{TableSegmAnalysis}
\centering
\caption{Performance analysis of BTS segmentation algorithm}\label{TableSegmAnalysis}
\resizebox{\textwidth}{!}{
\begin{tabular}{|l|c|p{7cm}|}
\hline
\textbf{Issue}                         & \textbf{Percentage and Number of BTS} & \textbf{Description}                                                                                           \\ \hline
Successful Segmentation                & \textbf{72.22\%}                       & BTS corpora accurately segmented without errors.                                                              \\ \hline
Inconsistencies in Numbering           & 19.43\%                       & Challenges due to numbering system inconsistencies, such as a section labeled as 4 followed by 1.1.            \\ \hline
Encoding Issues                        & 6.94\%                       & Section numbers appeared at the end of headlines after text extraction.                                         \\ \hline
Lack of Numbering System               & 1.38\%                      & Document lacked a numbering system altogether.                                                                 \\ \hline
\multicolumn{2}{|c|}{\multirow{2}{*}{\textbf{Overall Insight}}}               & An error-free numbering system is crucial for the algorithm's effective functionality.                         \\ \hline
\end{tabular}
}
\end{table}

\begin{algorithm}
\caption{Extract Hierarchical Sections}\label{alg2}
\KwData{path\_pdf: Path to the PDF file}
\KwResult{List of hierarchical sections}
 Initialize:\\
\hspace{15pt}ordered\_headings \tcp{List to store ordered headings}
\hspace{15pt}
 levels \tcp{Dictionary to track paragraph levels}
\hspace{15pt}
 regex \tcp{Regular expression pattern for headings}
\hspace{15pt}
 Lst \tcp{List to store hierarchical sections}
\hspace{15pt}
  root \tcp{Root node for the document hierarchy}
\hspace{15pt}
 previous\_nodes \tcp{Track previous nodes by level}
\hspace{15pt}
  node \tcp{Current node being processed}
\tcp{Extract each header along with its associated paragraph}

\ForEach{paragraph in the PDF}{
  \If{paragraph is not part of a header or footer}{
       Verify if the paragraph matches the regex pattern\;
       Extract the line text and level text\;
       Determine the paragraph's level based on numbering or indentation\;
       Ensure that the extracted line text and level text are valid\;
      
  }
  \If{paragraph is valid }{
       Set the current level\;
       Associate the paragraph text with the current node\;
       Append the paragraph to the node's list of paragraphs\;
  }
  \Else{
       Append the paragraph to the current node\;
      
  }}
\tcp{Merge paragraphs of sibling nodes into their parent nodes}
\ForEach{leaf node in the hierarchy}{
     Traverse the document hierarchy\;
     Combine paragraphs within the same section\;
     Store each section in the list Lst\;
    }
\Return{Lst}

\end{algorithm}

\subsection{Data Annotation}\label{DataAnnotation}
\begin{figure}
    \centering
    \includegraphics[width=\textwidth]{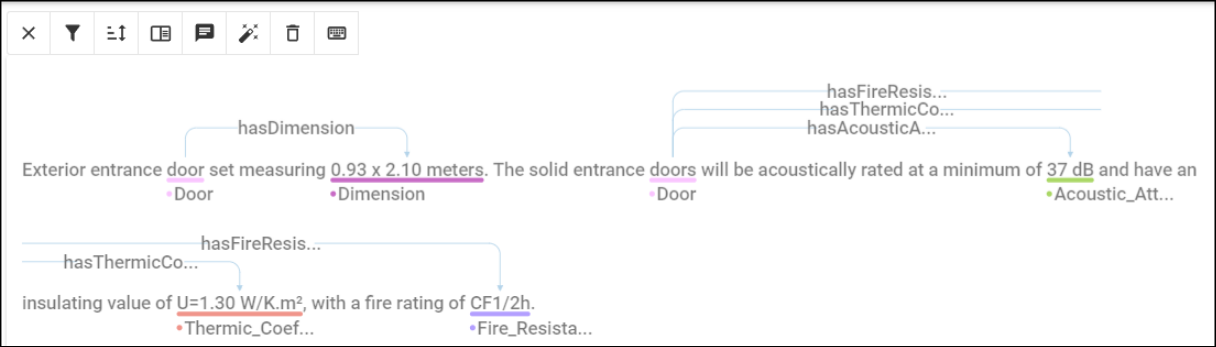}
    \caption*{Doccano interface displaying an example of text annotation.}  

    \begin{minipage}{0.5\textwidth}
        \resizebox{\textwidth}{!}{
        \begin{tabular}{lccc}
            \hline
            Id & Entity Label   & Offsets & Entity \\ \hline
            E1 & Door   & 18 22 & door \\  
            E2 & Dimension   & 37 55 & 0.93 x 2.10 meters \\
            E3 & Door  & 76 81 & doors \\
            ... & ...  & ... & ... \\
            E6 & Fire\_Resistance & 200 206 & CF1/2h \\ 
            \hline
        \end{tabular}
        }
        \caption*{Results from entity annotation}  
    \end{minipage}%
    \hfill  
    \begin{minipage}{0.5\textwidth}
        \resizebox{\textwidth}{!}{
        \begin{tabular}{lccc}
            \hline
            Id & From\_Id & To\_Id & Relation Label \\ \hline
            R1 & E1 & E2 & hasDimension \\
            ... & ... & ... & ... \\
            ... & ... & ... & ... \\
            R4 & E3 & E6 & hasFireResistance \\
            \hline
        \end{tabular}
        }
        \caption*{Results from relation annotation}  
    \end{minipage}
    \caption{Doccano's interface and annotation output example.}\label{fig:doccano_complete}
\end{figure}

Data annotation involves labeling concepts, properties, and relations in raw requirements, as shown in Figure \ref{fig:doccano_complete} using $Doccano$\footnote{Doccano: an open-source data labeling tool \url{https://doccano.github.io/doccano/}}. This process utilizes three BIM dictionaries: POBIM \cite{PlanBIM2022POBIM}, the Product Dictionary by the European Committee for Standardization \cite{Balaguer2017}, and the Model BIM Dictionary \cite{rapport_mission_numerique_batiment}. Manual analysis of the Building Technical Specifications enabled the selection of 233 pertinent raw requirements containing the defined concepts and properties from these dictionaries.

The use of these dictionaries addresses compatibility and standardization issues inherent in BIM data management across different software platforms. The Industry Foundation Classes (IFC) schema is designed to standardize BIM data but does not provide detailed properties on products performances like those mentioned in the BTS, which are essential for product specification, These dictionaries enhance standardized data handling, crucial for BIM model verification, and aid in developing systems that recommend products meeting specified requirements, for future research.

The output of the data annotation process, is formatted as JSON Lines files. Each line within these files adheres to a structured JSON format, encapsulating four essential elements: 1) Identifier (ID): Corresponds to the specific raw requirement. 2) Text Field: Contains the textual content of the raw requirement. 3) Entities Section: Holds the details of annotated entities, including their unique IDs, labels, and the text offsets which delineate where each entity appears within the raw text. 4) Relations Element: Defines the relationships among the annotated entities. Each relationship specifies an ID, the source (from\_id), and the target (to\_id), as well as the type of relationship involved. Figure \ref{fig:doccano_complete} demonstrates an example of annotated text as well as results obtained from entity and relation annotations.

\subsection{Data Splitting}
\begin{figure}[h!]
\centering
\includegraphics[width=\textwidth]{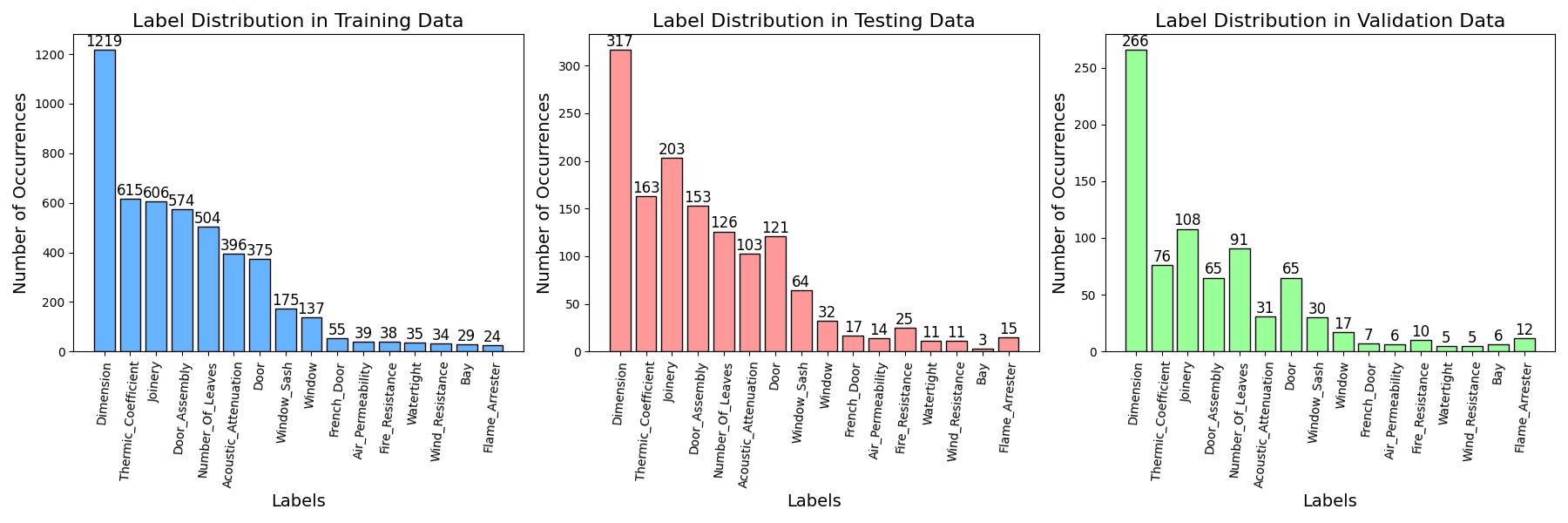}
\caption{Distribution of entity labels across training, testing, and validation datasets.}
\label{label_distribution}
\end{figure}

After labeling the data, it is necessary to split it. Data splitting is the partition of annotated data into three distinct datasets. 70\% of the data constitutes the training dataset to extract patterns and features from the data through manual or automatic methods. 20\% of the data is used for validation, that is, for tasks such as fine-tuning algorithm parameters and making adjustments. Finally, 10\% of the data is a testing dataset for evaluating algorithms. Figure \ref{label_distribution} shows the distribution of entity labels across training, testing, and validation data.

\subsection{NER for Concept and Property Extraction}
Based on the literature review (Section \ref{stateofart}), there are four NER approaches: rules, dictionaries, machine Learning, and deep learning. This study explores all four approaches.
\subsubsection{Rule-based and Dictionary-based Approaches for NER}
A manual analysis of the BTS showed that concepts primarily consist of entities, such as “Door”, “Window”, and “Window-sash”, which can be efficiently extracted using a predefined dictionary. In contrast, the properties exhibit a consistent structure, typically composed of numerical values paired with units, such as “m” for meters. Therefore, regular expressions and pattern matching were used for a rule-based approach. Table \ref{Table3} illustrates three distinct examples of properties and concepts, providing a visual representation of one possible form they could take.

\begin{table}[h]
\centering
\caption{Example of concepts and properties in joinery}\label{Table3}
\resizebox{\textwidth}{!}{
\begin{tabular}{|l|l|p{7cm}|}
\hline
\textbf{Concepts} & \textbf{Properties} & \textbf{Property Explanation} \\ \hline
Door (Porte) & 93x210 $cm$ & Dimension represents the physical size of the door, specified in centimeters (cm). \\ \hline
French door (Porte fenêtre) & R=1.46 W/m²°C & The thermal coefficient (R-value) indicating the insulation ability of the French door as an example. It may be expressed in Watts per square meter per degree Celsius (W/m²°C). \\ \hline
Window (Fenêtre) & RA,tr $\geq$ 35 dB & Describes the acoustic attenuation of the window for example, denoted as RA,tr, with a minimum requirement of 35 decibels (dB). \\ \hline
... & ... & ... \\ \hline
\end{tabular}
}
\end{table}

\paragraph{Regular Expressions for Concept and Property Extraction}
Information extraction using regular expressions is a technique that involves using predefined patterns or sequences of characters, known as regular expressions, to extract specific pieces of structured information from unstructured data. For instance, to extract the “Dimension” property “\(0.83 \times 2.19\) m” from a document, we can create a regular expression (pattern) that matches this specific format. In this case, the regular expression might be “\(\backslash d+\backslash.\backslash d+ \times \backslash d+\backslash.\backslash d+ m\)”. This regular expression is designed to find and extract numerical values in the format of decimal numbers followed by the character “\(\times\)” and another decimal number followed by the character “m”. By applying this regular expression to the text, we can effectively extract the value “\(0.83 \times 2.19\) m” of the property “Dimension”.

A manual analysis of \(70\%\) of our dataset (training dataset) enabled us to create a pattern that accounts for the varied expressions of each property that we tested with the testing set. A challenge to compare rule-based and dictionary-based approaches with machine learning-based approaches is using the same evaluation metrics. Precision (\(P\)), recall (\(R\)), and F1-score (\(F_1\)) are preferred to compare machine learning models \cite{zhang2021deep, moon2022automated, akanbi2021design}, so this paper follow a similar validation method for the rule-based approach (next Section \ref{validationprocess}). The extraction of concepts relies on merging the three dictionaries of BIM concepts/properties (previous Section \ref{DataAnnotation}) that serve as inputs to regular expressions that match the concepts in BTS.

\paragraph{Validation process}\label{validationprocess}
While NER tasks have traditionally employed rule-based approaches and manual or semi-automated evaluation \cite{guo2021semantic,ren2021semantic, wu2022rule}, this study sought a 6-step automated method to validate the rule-based approach (Figure \ref{validation}):

\begin{figure}[h!]
\centering
\includegraphics[width=\textwidth]{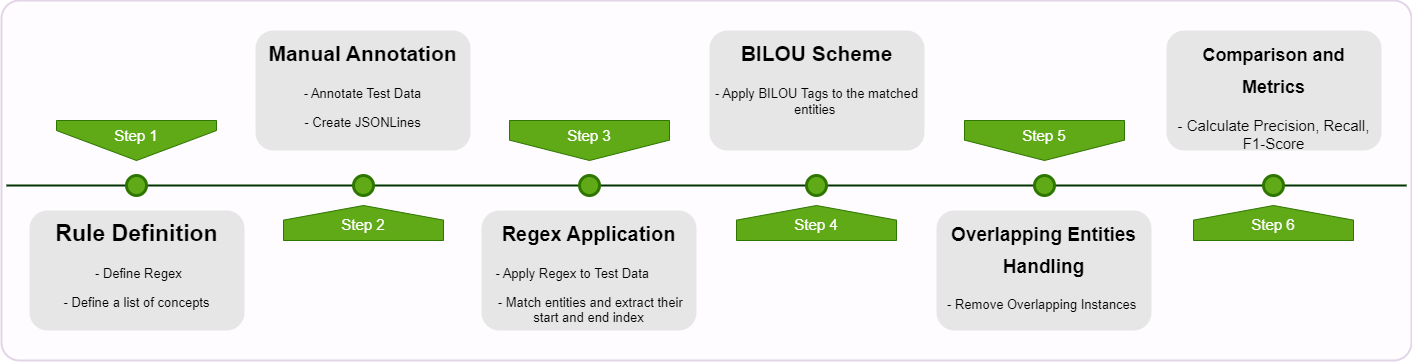}
\caption{Evaluation steps for our Rule-Based approach.}
\label{validation}
\end{figure}
Initially, a set of regular expressions to extract desired properties and concepts was defined. These regular expressions are constructed based on a manual training data analysis, and evaluated on the testing data set. The hand-crafted regular expressions extract both concepts and properties, providing not only the matched expressions but also their start, end offsets, and labels. These extracted details are then compared to the starting offset, ending offset, and labels found in our annotated dataset, Figure \ref{fig:doccano_complete}. This process enables the calculation of evaluation metrics.

\begin{figure}[h!]
\centering
\includegraphics[width=0.8\textwidth]{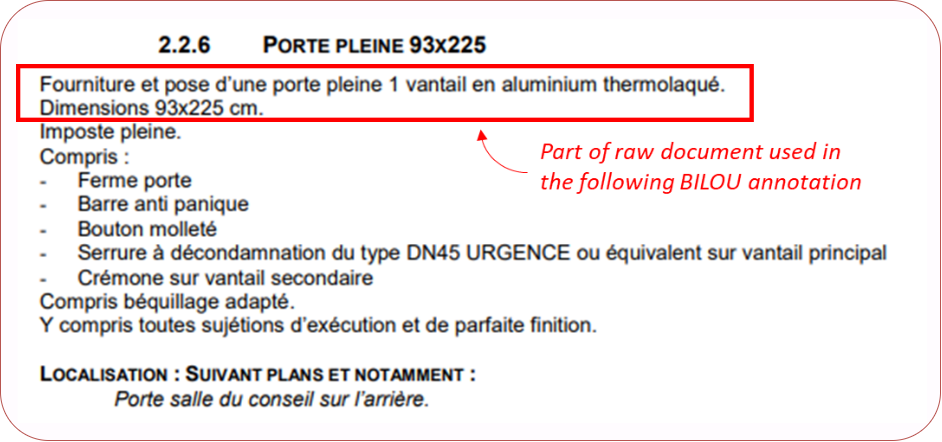}
\caption{Raw document part, where BILOU annotation will be done.}
\label{Performanceim}
\end{figure}

\subsubsection*{BILOU Scheme}

To further enhance entity boundary recognition and handling, the study use the Beginning, Inside, Last token, Outside, and Unit-length (BILOU) scheme \cite{alshammari2021impact,tkachenko2013named}. This scheme provides more detailed labeling, distinguishing the Beginning, Inside, Last token, and Unit-length chunks. BILOU is preferred for its ability to handle multi-word entities, compatibility with machine learning models, and detailed evaluation. Thus, given a raw requirement, the goal is to label each token \(x_i\) in the sentence \(X = (x_1, x_2, \ldots, x_n)\) with a BILOU tag scheme to obtain a tag sequence \(Y = (y_1, y_2, \ldots, y_n)\). Table \ref{Table4} shows a tag sequence for the phrase “\textit{fourniture et pose d’une porte pleine 1 vantail en aluminium thermolaqué. Dimensions 93×225 cm}” (Supply and installation of a 1-leaf solid door in thermolacquered aluminum. Dimensions 93×225 cm) in Figure \ref{Performanceim}. This example contains one concept “Door” (Porte), and two properties “$Number\_of\_Leaf$” (1 vantail), “$Dimension$” (93×225 cm). The tag sequence can decode the property/concept named entities, as seen in the last line of Table \ref{Table4}.

\begin{table}[ht]
\centering
\caption{An illustrative example of the tag sequence using the BILOU scheme.}\label{Table4}
\resizebox{\textwidth}{!}{
\begin{tabular}{|c|c|c|c|c|c|c|c|c|c|c|c|}
\hline
\textbf{Token} & Furniture & et & pose & d & ' & une & porte & pleine & 1 & vantail & en \\
\hline
\textbf{Tag sequence} & O & O & O & O & O & O & U-Door & O & B-Number\_of\_Leaf & L-Number\_of\_Leaf & O \\
\hline
\textbf{Entity type} & \multicolumn{6}{c|}{O} & Door &O & \multicolumn{2}{c|}{Number\_of\_Leaf} & O \\
\hline
\end{tabular}
}
\resizebox{0.8\textwidth}{!}{
\begin{tabular}{cc|c|c|c|c|c|}
\\
\hline
\multicolumn{1}{|c|}{\textbf{Token}} & \multicolumn{1}{c|}{aluminium} & thermolaqué & . & dimensions & 93x225 & cm \\
\hline
\multicolumn{1}{|c|}{\textbf{Tag sequence}} & O & O & O & O & B-Dimension & L-Dimension \\
\hline
\multicolumn{1}{|c|}{\textbf{Entity type}} & \multicolumn{4}{c|}{O} & \multicolumn{2}{c|}{Dimension}  \\
\hline
\end{tabular}
}
\end{table}
\subsubsection*{Overlapping Entities Handling}

Addressing overlapping entities involves extending the BILOU scheme to differentiate between primary and secondary categories of entities within overlapping text spans. In these spans, the BILOU tag is augmented with additional tags representing secondary entity types. For instance, a token might be part of a primary entity of type 'I-Dimension' and simultaneously part of a secondary entity of type 'B-Fire\_Resistance'. An extended tag such as 'I-Dimension\_B-Fire\_Resistance' resolves the issue by identifying entities that overlap within the span. Additionally, this extended tagging may be used to establish rules for prioritizing certain entity types over others during tagging. However, this study does not develop these rules, as that is beyond its scope.

\subsubsection*{Comparison and Metrics}

After applying the BILOU scheme to the annotated test data and the data matched by the rule-based method, a sequence of tokens was obtained \(X = (x_1, x_2,\ldots, x_n)\). Here, \(X\) is a list of tokens \(x_i\), comprising words, punctuation, etc. (as indicated in the line “token” of Table \ref{Table4}), collectively constructing the raw requirement. Simultaneously,  Simultaneously, the actual tag sequence \(Y_{\text{True}} = (y_1, y_2, \ldots, y_n)\) was determined, where \(Y_{\text{True}}\) comprises the true labels \(y_i\) manually labeled. The application of crafted regular expressions generated the predicted tag sequence \(Y_{\text{Predict}} = (y'_1, y'_2, \ldots, y'_n)\), where \(Y_{\text{Predict}}\) is a list of predicted labels \(y'_i\) associated with each token \(x_i\) in \(X\) the list. These tag sequences \(Y_{\text{True}}, Y_{\text{Predict}}\) are compared to calculate automatic evaluation metrics, including precision, recall, and F1-score. Additionally, a confusion matrix facilitates further analysis and insights into the method's performance. This validation process automatically assesses the performance of the rule-based approach compared with machine learning and deep-learning-based approaches.

\subsubsection{Machine Learning-Based Approaches for NER}

This section presents CRF models to extract the desired BIM concepts and properties, as they have demonstrated effectiveness in IE within technical domains\cite{skeppstedt2014automatic,patil2020named,lee2006fine,finkel2005incorporating}.

\paragraph{CRF for Concepts and Properties Extraction}

A set of features is constructed to equip a CRF model for NER. This ensemble of features encompasses linguistic and semantic elements, as outlined in Table \ref{Table5}, for each word in the raw requirement. Taking the example sentence "\textit{Porte vitrée à deux vantaux égaux}" (French door with two equal leaves),the list of features extracted for the word “$V\!antaux$” (Leaves) is illustrated in Table \ref{Table6}. Within this feature set, labels preceded by -1 or -2 refer to the previous word or the word before the previous relative to the target word. In the given example, the target word is '$vantaux$', and labels followed by +1 or +2 refer to the next word or the word after the next.

\begin{table}[ht]
\centering
\caption{Syntactic and Semantic features and their descriptions.}\label{Table5}
\resizebox{\textwidth}{!}{
\begin{tabular}{|l|l|p{6cm}|}
\hline
 & \textbf{Feature} & \textbf{Commentary} \\
\hline
\multirow{8}{*}{\textbf{Syntactic features}} & POS Tags (\texttt{postag}) & Assigning specific grammatical categories, such as noun, verb, or adjective, etc. \\
\cline{2-3}
 & Is digit (\texttt{word.isdigit()}) & Verify whether the word is a digit or not. \\
\cline{2-3}
 & Word length (\texttt{wordLength}) & Returns the number of characters in the word. \\
\cline{2-3}
 & Lowercase Word (\texttt{word.lower()}) & The lowercase version of the word. \\
\hline
\multirow{6}{*}{\textbf{Semantic features}} & \hspace{15pt}\multirow{6}{*}{5-word Window} & Considering a window of five words surrounding the target word (two words before and two words after), we associate the list of syntactic features with each word in the 5-word window. \\
\hline
\end{tabular}
}
\end{table}

\begin{table}[ht]
\centering
\caption{Example feature set for the word “vantaux” in a 5-Word window context.}\label{Table6}
\resizebox{0.95\textwidth}{!}{ 
\begin{tabular}{|p{5cm}|>{\arraybackslash}p{5cm}|>{\centering\arraybackslash}p{4cm}|}
\hline
\hspace{1cm}\textbf{Category} & \textbf{Feature} & \textbf{Value} \\ 
\hline
\hspace{0.5cm}\multirow{5}{*}{\textbf{Syntactic features}} 
 & \texttt{<word.lower()>} & “vantaux” \\
 & \texttt{<word.length()>} & 7 \\
 & \texttt{<word.isdigit()>} & False \\
 & \texttt{<postag>} & ``NOUN” \\
 \hline
 & \texttt{<-1.word.lower()>} & “deux” \\
 & \texttt{<-1.word.length()>} & 4 \\
 & \texttt{<-1.word.isdigit()>} & False \\
 & \texttt{<-1.postag>} & “NUM” \\
\hspace{0.5cm}\multirow{10}{*}{\textbf{Semantic features}} 
 & \texttt{<-2.word.lower()>} & “a” \\
 & \texttt{<-2.word.length()>} & 1 \\
 & \texttt{<-2.word.isdigit()>} & False \\
 & \texttt{<-2.postag>} & “VERB” \\
 & \texttt{<+1.word.lower()>} & “egaux” \\
 & \texttt{<+1.word.length()>} & 5 \\
 & \texttt{<+1.word.isdigit()>} & False \\
 & \texttt{<+1.postag>} & “ADJ” \\
 & \texttt{<+2.word.lower()>} & “\textbackslash n” \\
 & \texttt{<+2.word.length()>} & 1 \\
 & \texttt{<+2.word.isdigit()>} & False \\
 & \texttt{<+2.postag>} & “SPACE” \\
\hline
\end{tabular}}
\end{table}

The CRF model is trained on a labeled dataset, where text examples are paired with corresponding named entity labels. During the training phase, the model learns to recognize patterns and features within the input text that correlate with the presence of named entities. These patterns may encompass specific word occurrences, syntactic structures, or contextual cues. Once the CRF model has completed its training, it is primed for predicting named entities, assigning labels to each token based on its acquired patterns and features.
\subsubsection{Deep Learning-Based Approaches for NER}
The selection of the $BILSTM-CRF$ model was driven by its ability to process input sequences bidirectionally. The BILSTM architecture processes sequences both from the beginning to the end (forward pass) and from the end to the beginning (backward pass). Such dual processing facilitates the model's capacity to recognize dependencies and patterns over varying time steps. Within the BILSTM network, each LSTM unit retains a memory cell capable of storing and accessing information across extended sequences, which is crucial for managing long-range dependencies. The addition of the CRF layer enhances prediction accuracy by taking into account label dependencies and ensuring label consistency, rendering the BILSTM-CRF model particularly effective for NER tasks.

Transfer learning was also implemented, with a primary focus on addressing the labor-intensive and time-consuming process of annotating extensive datasets. This approach also served to evaluate the effectiveness of transfer learning within a domain-specific framework. The model $Fr\_core\_news\_lg$, a French large model grounded in convolutional neural network architecture and trained on extensive French datasets, was employed for this purpose.

Furthermore, this paper explores LLMs, particularly MLMs such as $CamemBERT$, which is based on the RoBERTa architecture \cite{liu2019roberta}. The objective is to fine-tune these models and evaluate their efficacy in entity recognition within specialized fields, with a focus on the French construction industry, particularly in BTS documents. A comparative analysis of $CamemBERT$ and $Fr\_core\_news\_lg$, along with other approaches developed in this study, provided valuable insights into their respective strengths for specialized NER tasks.

\subsection{Classification for Relation Extraction}
The supervised approach with feature-based methods was selected to develop the relation extraction model, driven by its established efficacy in domain-specific relation extraction tasks \cite{alimova2020multiple}. Supervised methods excel at learning from labeled data, which facilitates the recognition of complex patterns within relations. Additionally, these methods provide flexibility through customizable features, enabling the adaptation of the model to specific domain and relation types.

Figure \ref{fig:doccano_complete} illustrates that the training involves a dataset containing relations extracted from annotated data. Various syntactic and semantic features, as detailed in \cite{bach2007review}, were utilized to identify the relationship between entities within a given raw requirement. Moreover, another feature detailed in \cite{alimova2020multiple} was incorporated.  Table \ref{sandsy} contains the list of features utilized in the analysis.
\begin{longtable}{|p{\dimexpr0.16\textwidth-2\tabcolsep-\arrayrulewidth\relax}| 
                  p{\dimexpr0.09\textwidth-2\tabcolsep-\arrayrulewidth\relax}| 
                  p{\dimexpr0.28\textwidth-2\tabcolsep-\arrayrulewidth\relax}| 
                  p{\dimexpr0.47\textwidth-2\tabcolsep-\arrayrulewidth\relax}|} 
\caption{Semantic (Sem) and Syntactic (Syn) features for entity relationship analysis in raw requirements.}\label{sandsy}  \\
\hline
\textbf{Feature} & \textbf{Type} & \textbf{Definition} & \textbf{Example} \\ \hline

Entities' Span & Sem & The extent or range of words covered by each entity in the sentence. & In “glazed door with dimensions 100×220,” the span of the entities \textit{“Door”} and \textit{“Dimension”} are “door”, “100×220” successively. For each entity, we apply \textit{Word2Vec} and then calculate the average of the entity vectors if it is composed of multiple words. \\ \hline

Labels Assigned to Entities & Sem & The semantic category or type assigned to each entity, indicating its role or nature. & In our NER task, the label for “100×220” is  “\textit{Dimension}”. \\ \hline

Entities' POS tags & Syn & The POS of the two entities, capturing their syntactic context. & In “glazed door with dimensions 100×220.”, the POS tags of entities: “door” is a (NOUN) noun, and “100×220” is a (PROPN) proper noun. \\ \hline

The sequence of Words & Sem & The sequential arrangement of the average of word embedding between two entities, capturing their semantic context. & In “glazed door with dimensions 100×220”, the sequence of words between “door” and “100×220” is “with dimensions”. In this case, we use the Word2Vec model for “with” and “dimension”, and then calculate the average of the word vectors. \\ \hline

Count of Words & Syn & The number of words in the sequence between two entities, providing a quantitative measure of syntactic distance. & In “glazed door with dimensions 100×220.”,  the count of words between “door” and “100×220” is 2. \\ \hline

Path Within Parse Tree & Sem & The route or series of syntactic relationships connecting two entities in the parse tree of a sentence. & In a parse tree, the path between “door” and “100×220” is “nmod (nominal modifier) → nummod (numeric modifier). \\ \hline

Number of Sentences & Syn & The count of sentences between entities. & In the same example, the number of sentences between “door” and “100×220” is 0. \\ \hline

Punctuation Characters & Syn & The count of punctuation characters between entities. & The punctuation character count between “door” and “100×220” is 0. \\ \hline

Orientation & Syn & Specifies the relative positioning of entity 1 in relation to entity 2, indicating whether entity 1 comes before or after entity 2. & The orientation of “door” to “100×220” in the same example is “before”. \\ \hline

Title of Raw Requirement & Sem & The title of the raw requirement section. & The paragraph P4 in Figure \ref{ILPAR} is titled “Porte vitrée 100×220” (glazed door 100×220), so the title is “PORTE VITREE 100×220”. For each word, we apply \textit{Word2Vec}, and then we calculate the average of the word vectors. \\ \hline

\end{longtable}

A manual examination showed that, in many cases, if a concept is presented in the title, Figure \ref{ILPAR} (P1, P2, P3, P4, P6), it represents the primary concept of the raw requirement, and this is the concept from which the associated properties must be extracted. The Word2Vec model with Skip-gram architecture \cite{rong2014word2vec} was employed  with a vector size of 300 to vectorize each word comprising the title of the raw requirement or entities themselves. The average of the vectors constituting the words in the title or entities was computed to generate the sentence vector (sentence embedding feature).

The relevance of these features is determined through a Recursive Feature Elimination (RFE)\footnote{Recursive Feature Elimination (RFE): \url{https://scikit-learn.org/stable/modules/generated/sklearn.feature_selection.RFE.html}} method. To demonstrate the results of the best combination of features, Seven combinations of features were created to extract the best combination of features for the current task. This is illustrated in Figure \ref{fig18} using the Random Forest classifier. The feature combinations and their impact on the performance metrics found in Table \ref{tab:feature_combos}

\begin{table}[h]
\centering
\caption{Impact of Feature Combinations on Relation Extraction Model}\label{tab:feature_combos}
\resizebox{0.95\textwidth}{!}{ 
\begin{tabular}{|c|p{6cm}|c|}
\hline
\textbf{Combination} & \textbf{Features Added} & \textbf{Performance (F1-score)} \\ \hline
1 & Labels Assigned to Entities, Entities' POS tags & $\leq$ 0.4 for relations like ``hasDimension'' \\ \hline
2 & Addition to comb1: Count of Words, Number of Sentences, Punctuation Characters & $\geq$ 0.7 for relations like ``hasNumberOfLeaf'' \\ \hline
3 & Addition to comb2: Orientation & $\geq$ 0.8 for relations like ``hasAcousticAttenuation'' \\ \hline
4 & Addition to comb3: Path Within Parse Tree & Decreased to 0.60 for ``hasDimension'' \\ \hline
5 & Addition to comb3: Sequence of Words & Decreased to 0.25 for ``hasDimension'' \\ \hline
6 & Addition to comb3: Entities' Span & Decreased to 0.45 for ``hasFireResistance'' \\ \hline
7 & Addition to comb3: Title of Raw Requirement & No significant change \\ \hline
\end{tabular}
}
\end{table}

From Table \ref{tab:feature_combos}, it is concluded that in this case, semantic features negatively impact the results, suggesting that the entities depend more on syntactic than semantic features.

In conclusion, after examining Figure \ref{fig18}, it is observed that for combinations 2 to 7, the classifier achieves high precision, effectively minimizing false positives. However, variations in recall across the combination sets suggest that the features influence the classifier's ability to mitigate false negatives. Analyzing the F1-score, which is the harmonic mean of precision and recall, combination 3, which includes features such as \textit{Labels Assigned to Entities + Entities' POS tags + Count of Words + Number of Sentences + Punctuation Characters + Orientation}, provides the best results across all relations, thus becoming the chosen feature combination.

\begin{figure}[h!]
\centering
\includegraphics[width=\textwidth]{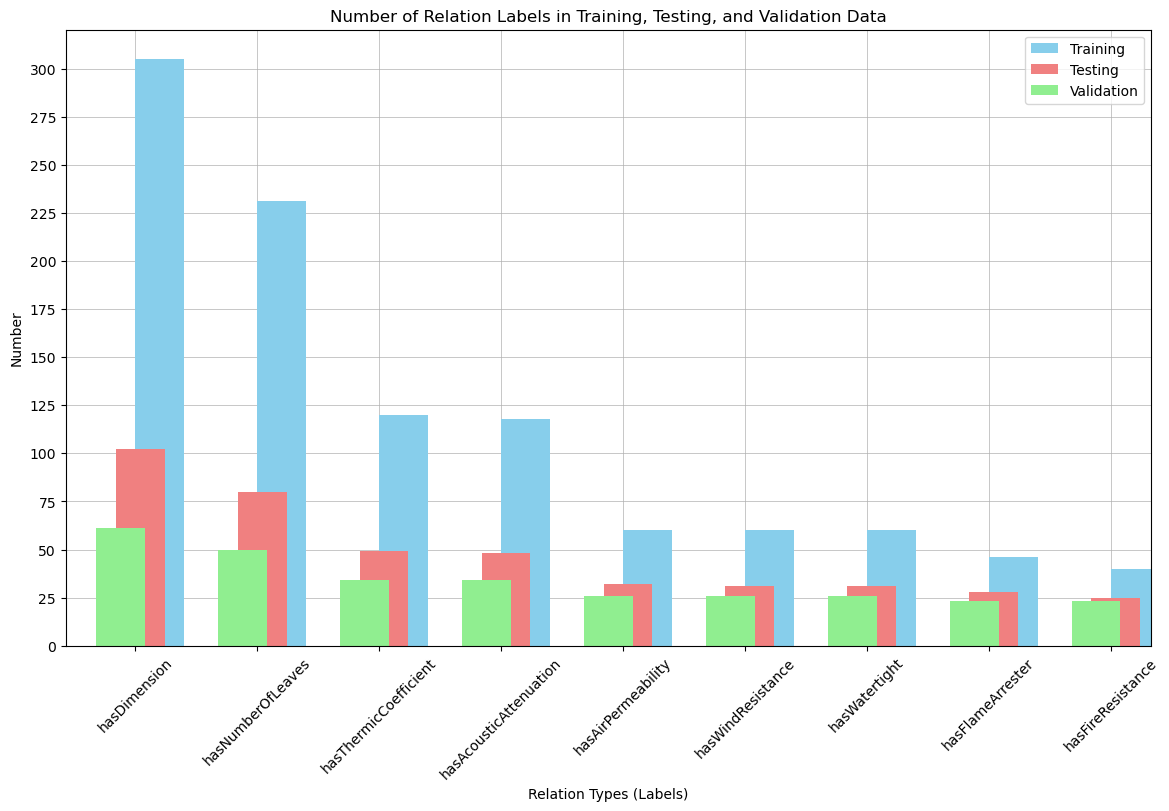}
\caption{Number of relation labels in training, testing and validation data.}
\label{Performanceim2}
\end{figure}

\begin{figure}[h!]
\centering
\includegraphics[width=\textwidth]{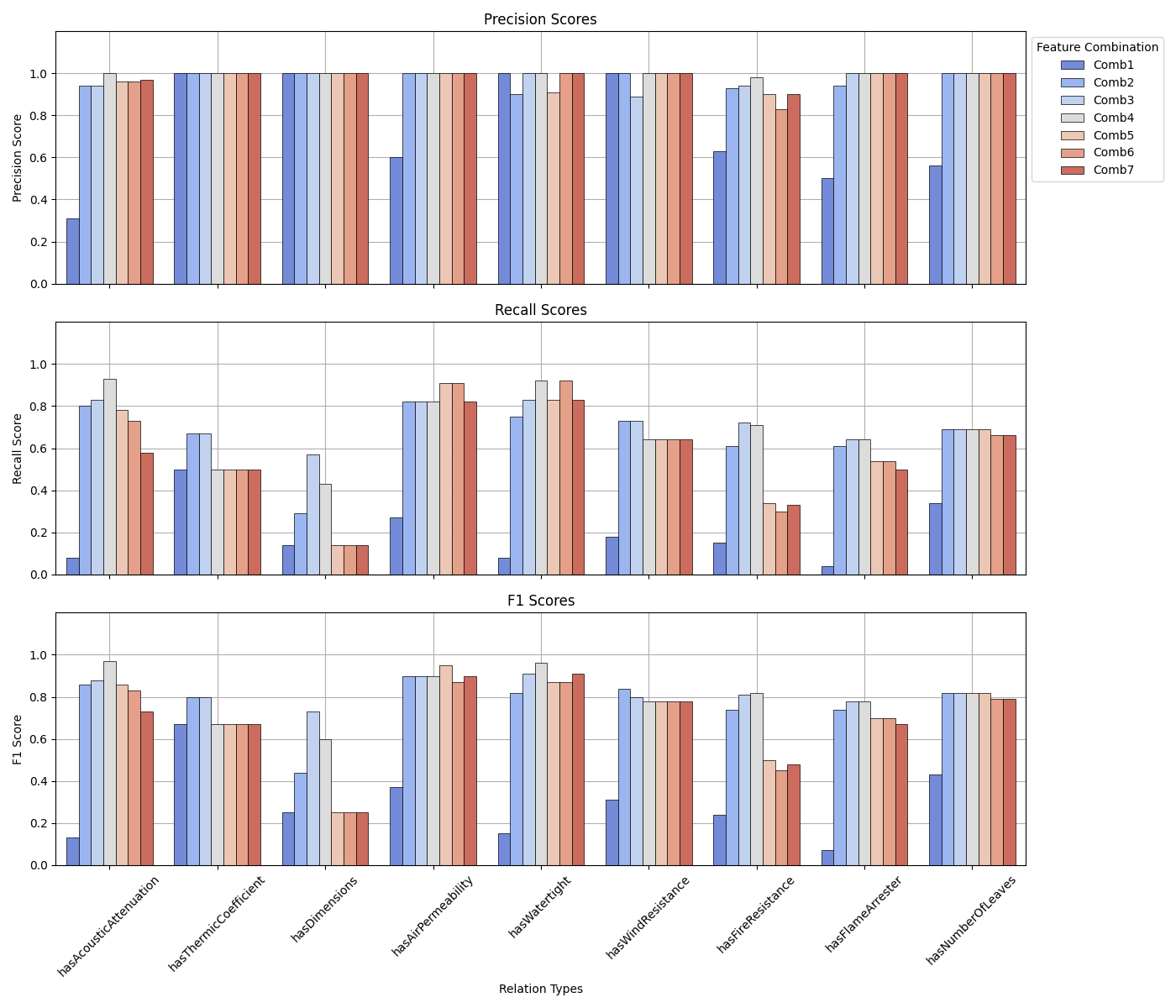}
\caption{Performance comparison of feature combinations for relationship extraction.}
\label{fig18}
\end{figure}

The dataset was partitioned into training (70\%), testing (20\%), and validation (10\%) for the RE classification task. Figure \ref{Performanceim2} displays the distribution of each relation type in the training, testing and validation corpora, excluding the “0” label, which indicates no relation between entities. Four classifiers, SVM, RF, DT, and KNN, were benchmarked for relation extraction. These models, known to perform well in RE tasks \cite{alimova2020multiple}, \cite{peng2018chemical}, \cite{zhao2023biomedical}, were selected for their effectiveness with small annotated datasets. The ability to design custom feature vectors lends these models the flexibility required to capture nuanced relationships between entities, positioning them as an optimal starting point for this study.

\begin{figure}[h!]
\centering
\includegraphics[width=\textwidth]{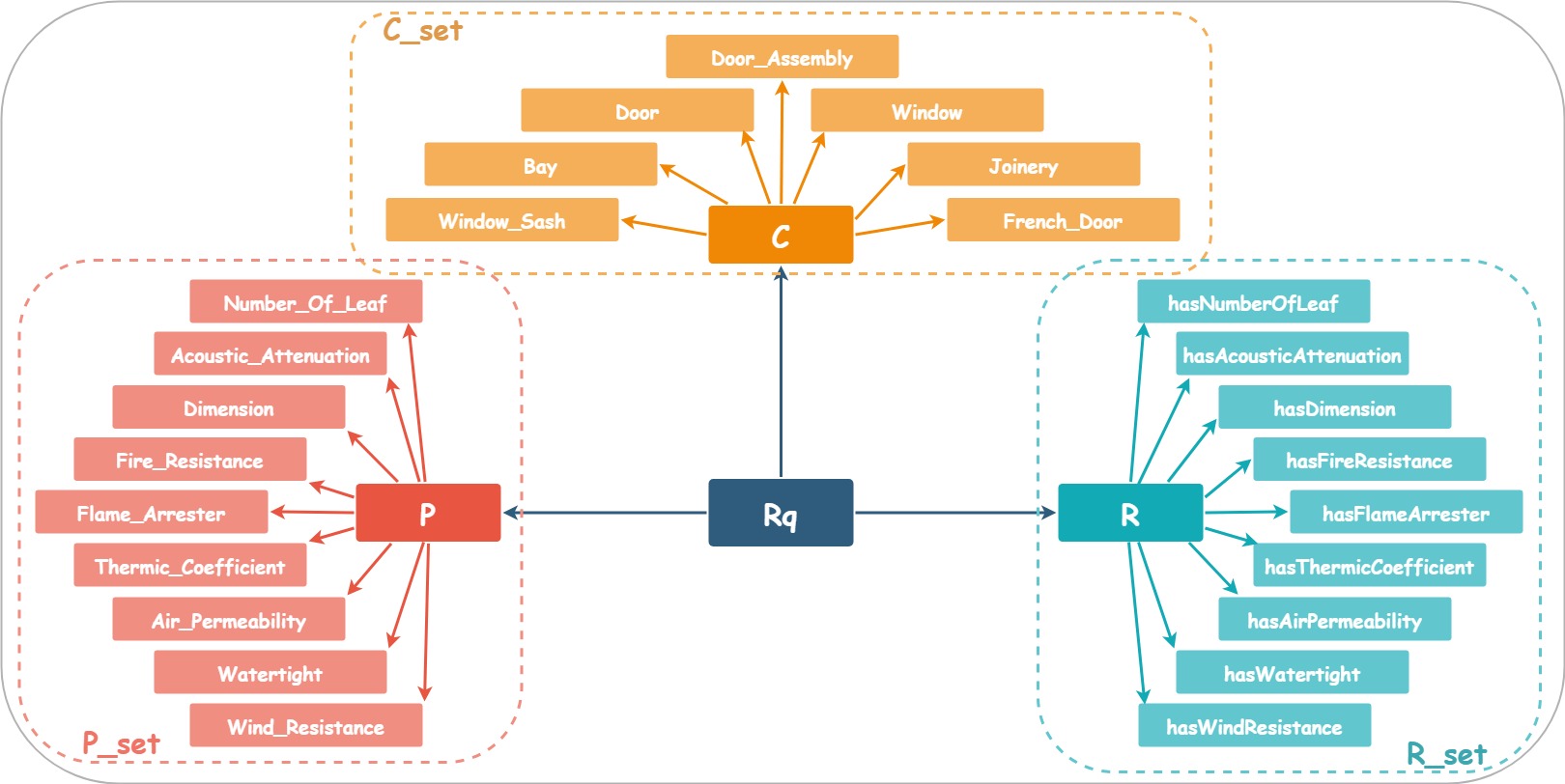}
\caption{Aimed entities and relations in joinery BTS documents.}
\label{fig19}
\end{figure}

\section{Validation: Benchmarking on Joinery BTS}\label{exp}

This section is a benchmark of the NER and RE models presented in Section \ref{methode}, including selecting hyperparameters and fine-tuning key parameters to maximize the models' accuracy and generalization to real-world data. The entities and the relations that this study aim to extract are the concepts and properties of joinery BTS (Figure \ref{fig19}).

\subsection{NER Experiments}
Table \ref{tab:ner_config} outlines the specific configurations used for each model in the experiment step. Throughout the experimental phases, the callback API\footnote{Keras documentation: Callbacks API: \url{https://keras.io/api/callbacks/}} was employed to monitor and adjust the training process dynamically.

\begin{table}[h]
\centering
\caption{Experiment configuration and hyperparameter Tuning for NER task}
\label{tab:ner_config}
\resizebox{0.95\textwidth}{!}{ 
\begin{tabular}{|l|p{10cm}|}
\hline
\textbf{Model} & \textbf{Configuration} \\
\hline
CRF  & Uses Limited-memory Broyden-Fletcher-Goldfarb-Shanno (BFGS) to fine-tune the hyperparameters. Its advantages include efficient memory usage and fast convergence suited for large-scale structured prediction. Hyperparameters \( c_1 \) and \( c_2 \) set to 0.1 with a maximum iteration limit of 100. \\
\hline
BiLSTM-CRF & Embedding dimension of 128 and 64 hidden units to facilitate learning. \\
\hline
Camembert\_base & Pre-trained transformer with a batch size of 32 and learning rate \( 2 \times 10^{-5} \). \\
\hline
Fr\_core\_news\_lg & Learning rate managed by default by spaCy. \\
\hline
\end{tabular}
}
\end{table}

\subsubsection{NER Results}
\begin{table*}[h!]
\centering
\caption{a, b, c, d: Comparative results of models on different entities. The highest precision (P), recall (R), and F1-score (F1) are in bold.}\label{table8}
\resizebox{\textwidth}{!}{
\begin{tabular}{lccc|ccc|ccc|ccc} 
\\
\multicolumn{13}{l}{Table 8.a: Comprehensive performance of the NER models on the test corpus.}
\\
\hline
\multirow{3}{*}{\textbf{Models}} & \multicolumn{12}{c}{Entities} \\
\cline{2-13}
                                 & \multicolumn{3}{c}{Door\_Assembly} & \multicolumn{3}{c}{Bay} & \multicolumn{3}{c}{Window} & \multicolumn{3}{c}{Joinery} \\ 
\cline{2-13}
                                 & P & R & F1 & P & R & F1 & P & R & F1 & P & R & F1 \\
\hline
Rule\_based                      & \textbf{1.00} & \textbf{1.00} & \textbf{1.00} & \textbf{0.75} & \textbf{1.00} & \textbf{0.86} & \textbf{0.97} & 0.94 & 0.95 & 0.96 & 0.74 & 0.84 \\
CRF                              & \textbf{1.00} & 0.98 & 0.99 & 0.67 & 0.67 & 0.67 & \textbf{0.97} & \textbf{1.00} & \textbf{0.98} & \textbf{0.96} & \textbf{1.00} & \textbf{0.98} \\
BiLSTM-CRF                       & 0.95 & 0.97 & 0.96 & 0.43 & \textbf{1.00} & 0.60 & 0.94 & \textbf{1.00} & 0.97 & 0.95 & 0.98 & 0.97 \\
$Camembert\_base$                  & 0.99 & 0.93 & 0.96 & \textbf{0.75} & \textbf{1.00} & \textbf{0.86} & 0.71 & \textbf{1.00} & 0.83 & 0.91 & \textbf{1.00} & 0.95 \\
$Fr\_core\_news\_lg$               & 0.99 & \textbf{1.00} & 0.99 & \textbf{0.75} & \textbf{1.00} & \textbf{0.86} & 0.97 & \textbf{1.00} & 0.98 & 0.89 & \textbf{1.00} & 0.94 \\
\hline
\\
\multicolumn{13}{l}{Table 8.b: Comprehensive performance of the NER models on the test corpus.}
\\
\hline

\multirow{3}{*}{\textbf{Models}} & \multicolumn{12}{c}{Entities} \\
\cline{2-13}
                                 & \multicolumn{3}{c}{Window\_Sash} & \multicolumn{3}{c}{Door} & \multicolumn{3}{c}{French\_Door} & \multicolumn{3}{c}{Number\_of\_Leaf} \\ 
\cline{2-13}
                                 & P & R & F1 & P & R & F1 & P & R & F1 & P & R & F1 \\
\hline
Rule\_based                      & \textbf{1.00} & \textbf{1.00} & \textbf{1.00} & 0.79 & \textbf{0.99} & 0.88 & \textbf{1.00} & 0.94 & 0.97 & \textbf{1.00} & 0.90 & 0.95 \\
CRF                              & \textbf{1.00} & \textbf{1.00} & \textbf{1.00} & \textbf{0.96} & \textbf{0.99} & \textbf{0.98} & \textbf{1.00} & \textbf{1.00} & \textbf{1.00} & \textbf{1.00} & 0.98 & 0.99 \\
BiLSTM-CRF                       & \textbf{1.00} & \textbf{1.00} & \textbf{1.00} & 0.94 & 0.98 & 0.96 & 0.50 & 0.50 & 0.50 & 0.87 & 0.95 & 0.91 \\
$Camembert\_base $                 & \textbf{1.00} & \textbf{1.00} & \textbf{1.00} & 0.95 & \textbf{0.99} & 0.97 & \textbf{1.00} & \textbf{1.00} & \textbf{1.00} & 0.97 & 0.98 & 0.98 \\
$Fr\_core\_news\_lg$               & \textbf{1.00} & 0.98 & 0.99 & 0.95 & \textbf{0.99} & 0.97 & 0.94 & \textbf{1.00} & 0.97 & \textbf{1.00} & \textbf{1.00} & \textbf{1.00} \\
\hline
\\
\multicolumn{13}{l}{Table 8.c: Comprehensive performance of the NER models on the test corpus.}
\\
\hline

\multirow{3}{*}{\textbf{Models}} & \multicolumn{12}{c}{Entities} \\
\cline{2-13}
                                 & \multicolumn{3}{c}{Acoustic\_Attenuation} & \multicolumn{3}{c}{Dimension} & \multicolumn{3}{c}{Fire\_Resistance} & \multicolumn{3}{c}{Flame\_Arrester} \\ 
\cline{2-13}
                                 & P & R & F1 & P & R & F1 & P & R & F1 & P & R & F1 \\
\hline
Rule\_based                      & 0.93 & \textbf{1.00} & 0.96 & 0.76 & 0.84 & 0.80 & \textbf{1.00} & \textbf{1.00} & \textbf{1.00} & \textbf{1.00} & 0.60 & 0.75 \\
CRF                              & \textbf{1.00} & 0.97 & 0.99 & 0.97 & 0.94 & 0.96 & 0.70 & 0.56 & 0.62 & \textbf{1.00} & 0.40 & 0.57 \\
BiLSTM-CRF                       & 0.93 & 0.90 & 0.92 & 0.68 & 0.61 & 0.64 & 0.43 & 0.55 & 0.48 & \textbf{1.00} & 0.57 & 0.73 \\
$Camembert\_base$                  & 0.98 & 0.98 & 0.98 & 0.98 & 0.97 & 0.98 & \textbf{1.00} & \textbf{1.00} & \textbf{1.00} & 0.86 & 0.80 & 0.83 \\
$Fr\_core\_news\_lg$               & 0.97 & \textbf{1.00} & \textbf{0.99} & \textbf{0.98} & \textbf{0.98} & \textbf{0.98} & \textbf{1.00} & 0.96 & 0.98 & 0.88 & \textbf{0.93} & \textbf{0.90} \\
\hline
\\
\multicolumn{13}{l}{Table 8.d: Comprehensive performance of the NER models on the test corpus.}
\\
\hline

\multirow{3}{*}{\textbf{Models}} & \multicolumn{12}{c}{Entities} \\
\cline{2-13}
                                 & \multicolumn{3}{c}{Thermic\_Coefficient} & \multicolumn{3}{c}{Air\_Permeability} & \multicolumn{3}{c}{Watertight} & \multicolumn{3}{c}{Wind\_Resistance} \\ 
\cline{2-13}
                                 & P & R & F1 & P & R & F1 & P & R & F1 & P & R & F1 \\
\hline
Rule\_based                      & \textbf{1.00} & 0.79 & 0.88 & 0.93 & \textbf{1.00} & \textbf{0.97} & 0.37 & \textbf{1.00} & 0.54 & \textbf{1.00} & \textbf{1.00} & \textbf{1.00} \\
CRF                              & 0.99 & 0.96 & 0.98 & \textbf{1.00} & 0.79 & 0.88 & \textbf{1.00} & 0.82 & 0.90 & \textbf{1.00} & 0.82 & 0.90 \\
BiLSTM-CRF                       & 0.86 & 0.86 & 0.86 & \textbf{1.00} & 0.79 & 0.88 & \textbf{1.00} & 0.91 & 0.95 & \textbf{1.00} & 0.91 & 0.95 \\
$Camembert\_base$                 & \textbf{1.00} & 0.96 & 0.98 & \textbf{1.00} & 0.86 & 0.92 & 0.91 & 0.91 & 0.91 & \textbf{1.00} & \textbf{1.00} & \textbf{1.00} \\
$Fr\_core\_news\_lg$               & 0.99 & \textbf{0.99} & \textbf{0.99} & \textbf{1.00} & 0.93 & 0.96 & \textbf{1.00} & 0.91 & 0.95 & 0.92 & \textbf{1.00} & 0.96 \\
\hline
\end{tabular}
}
\end{table*}

Tables \ref{table8}. a, \ref{table8}. b, \ref{table8}. c, and \ref{table8}.d show the results of evaluating NER models on various properties.

\subsubsection{Discussion}
The \textbf{Rule-based model} demonstrates robust performance with high average F1-scores (0.92 for concept extraction and 0.87 for property extraction), indicating strong precision and recall. However, in some cases, the rule-based model demonstrates lower recall due to its reliance on predefined patterns. In the test data, new patterns were discovered that were not present in the training data. This indicates a failure to account for these patterns when creating regular expressions, resulting in their omission during extraction. This oversight increases the number of false negatives, consequently affecting the recall value. It is essential to acknowledge the limitations of the rule-based approach, which requires updating rules when new patterns or rules are discovered. On the other hand, lower precision may result in an increased number of false positives being extracted. For example, consider the entity “$Door$”, translated to “$Porte$” in French. “$Porte$” is also a conjugated form in the present tense of the verb “$Porter$”, meaning “$To\ carry$”, This can lead to extraction errors and suggests additional post-processing steps, such as filtering based on values.

The \textbf{CRF model} also performs well, achieving an F1-score of 0.94 for concept extraction and 0.86 for property extraction. but may vary depending on the entity. Entities with limited representation in the training data, such as “Bay”, “$Fire\_Resistance$”, and “$Flame\_Arrester$”, can impact the model's ability to generalize effectively to the test data. The scarcity of examples for such entities poses a challenge for the model to capture robust patterns and may result in variations in its performance across different entity types. 

The \textbf{BiLSTM-CRF model}, with an F1-score of 0.81 for concept extraction and 0.85 for property extraction, faces challenges with lower F1-scores for specific entities compared to other approaches. Notably, some entities pose greater difficulties for the BiLSTM-CRF model in terms of generalization. For instance, “$French\_Door$” achieves an F1-score of 0.50, “$Fire\_Resistance$” 0.48, and both “Bay” and “Dimension” achieve F1-scores of 0.60. Upon reviewing our training data, it becomes evident that certain entities such as “Bay” and “French door” (as illustrated in Figure \ref{label_distribution}) have significantly fewer instances compared to others. Additionally, the complexity of patterns associated with the “Dimension” property presents challenges for the model's generalization. 

The transfer learning approach yielded an F1-score of 0.95 for concept extraction and 0.96 for property extraction for \textbf{Fr\_core\_news\_lg}. Similarly, \textbf{Camembert\_base} achieved an F1-score of 0.93 for concept extraction and 0.95 for property extraction. It is evident from the results that these two models perform well even with entities that present challenges for other models, such as "Flame\_Arrester". This highlights the effectiveness of such methods, which alleviate the effort required to annotate a large dataset or create handcrafted rules. While the study demonstrates that handcrafted rules can yield good results, they necessitate ongoing maintenance and adaptation, as observed also in this study.

\subsection{RE Experiments}
After extracting entities, the next step was determining their relationships (Figure \ref{fig19}), which is essential for building a formalized requirement, as previously mentioned. We employed supervised machine learning using four classifiers: SVM, RF, DT, and KNN. These relationships exist between a Concept ($\mathcal{C}$) and a property ($\mathcal{P}$), with no relationships between concepts or properties.
\subsubsection{Hyperparameter Tuning for Supervised Machine Learning RE Models}
To select optimal hyperparameters for the supervised machine learning models, including SVM, RF, DT, and KNN, we employed a grid search. Grid search allows us to explore a wide range of hyperparameters by systematically testing various combinations, ultimately identifying the most suitable settings.

Table \ref{tab:hyperparameter_tuning} provides details of the hyperparameter tuning process for the four models. For each model, the optimal hyperparameter settings, explored parameter values, and their definitions are presented.

\begin{table}[h]
\caption{Hyperparameter tuning details for machine learning models}
\label{tab:hyperparameter_tuning}
\resizebox{\textwidth}{!}{
\begin{tabular}{|l|l|p{2.5cm}|p{5cm}|p{7cm}|}
\hline
\textbf{Model} & \textbf{Parameter} & \textbf{Optimal Value} & \textbf{Explored Values} & \textbf{Definition} \\
\hline
\multirow{2}{*}{SVM} & $C$ & 1.0 & 0.1, 1, 10 & Regularization parameter controlling the trade-off between achieving a low training error and a low testing error \\
\cline{2-5}
 & $\gamma$ & "auto" & 0.001, 0.01, 0.1, "auto", "scale" & Kernel coefficient \\
\hline
\multirow{3}{*}{RF} & $n\_estimators$ & 200 & 50, 100, 200 & Number of trees in the forest \\
\cline{2-5}
 & $max\_depth$ & 30 & 10, 20, 30 & Maximum depth of each tree \\
\cline{2-5}
 & $min\_samples\_split$ & 2 & 2, 5, 10 & Minimum samples required for splitting nodes \\
\hline
\multirow{4}{*}{DT} & $Criterion$ & "entropy" & "gini", "entropy" & Function to measure the quality of a split \\
\cline{2-5}
 & $max\_depth$ & 30 & 10, 20, 30 & Maximum depth of each tree \\
\cline{2-5}
 & $min\_samples\_split$ & 2 & 2, 5, 10 & Minimum samples required for splitting nodes \\
\cline{2-5}
 & $max\_features$ & "sqrt" & "sqrt", "log2", "auto" & Maximum number of features to consider for the best split \\
\hline
\multirow{3}{*}{KNN} & $n\_neighbors$ & 5 & 3, 5, 7, 9, 10, 13, 15 & Number of neighbors to use for kneighbors queries \\
\cline{2-5}
 & Distance Metric & "Manhattan" & "Euclidean", "Manhattan", "Minkowski" & Distance metric used \\
\cline{2-5}
 & Weight & "distance" & "uniform", "distance" & Weight function used \\
\hline
\end{tabular}
}
\end{table}

\subsubsection{RE Results}
The metrics used for evaluating and comparing The RE models are similar to those used in NER models. 
Table \ref{table9}. (a, b) presents the results of each model using syntactic features, including precision, recall, and F1-score for each property.
Table \ref{table10}. (a, b) presents the results of each model using syntactic and semantic features.
Table \ref{table11} presents a comprehensive performance analysis of each model for extracting all relations using syntactic features and then using both syntactic and semantic features.

\begin{table*}[h!]
\centering
\caption{Comprehensive performance of the RE models on the test corpus using syntactic features. The highest precision (P), recall (R), and F1-score (F1) are in bold.}\label{table9}
\resizebox{0.88\textwidth}{!}{
\begin{tabular}{lccc|ccc|ccc|ccc} 
\\
\multicolumn{13}{l}{Table 9. a: Comprehensive performance of the RE models on the test corpus using syntactic features.} \\
\hline
\multirow{3}{*}{\textbf{Models}} & \multicolumn{12}{c}{Relations} \\
\cline{2-13}
                                 & \multicolumn{3}{c}{$hasDimension$} & \multicolumn{3}{c}{$hasThermicCoe\!f\!ficient$} & \multicolumn{3}{c}{$hasAcousticAttenuation$} & \multicolumn{3}{c}{$hasFlameArrester$} \\ 
\cline{2-13}
                                 & P & R & F1 & P & R & F1 & P & R & F1 & P & R & F1 \\
\hline
SVM                              & 0.88 & 0.44 & 0.59 & 0.93 & 0.50 & 0.65 & \textbf{1.00} & 0.69 & \textbf{0.82} & \textbf{1.00} & 0.50 & 0.67 \\
RF                               & \textbf{0.92} & 0.74 & \textbf{0.82} & \textbf{1.00} & 0.64 & \textbf{0.78} & \textbf{1.00} & 0.69 & \textbf{0.82} & \textbf{1.00} & \textbf{0.67} & \textbf{0.80} \\
DT                               & 0.75 & \textbf{0.84} & 0.79 & 0.68 & \textbf{0.75} & 0.71 & 0.81 & \textbf{0.72} & 0.76 & 0.80 & \textbf{0.67} & 0.73 \\
KNN                              & 0.89 & 0.67 & 0.76 & 0.94 & 0.57 & 0.71 & 0.91 & 0.69 & 0.78 & 0.80 & \textbf{0.67} & 0.73 \\
\hline
\end{tabular}}
\resizebox{\textwidth}{!}{
\begin{tabular}{lccc|ccc|ccc|ccc|ccc} 
\\
\multicolumn{16}{l}{Table 9. b: Comprehensive performance of the RE models on the test corpus using syntactic features.} \\
\hline
\multirow{3}{*}{\textbf{Models}} & \multicolumn{12}{c}{Relations}
\\ \cline{2-16}
& \multicolumn{3}{c|}{\textbf{$hasFireResistance$}} & \multicolumn{3}{c|}{\textbf{$hasAirPermeability$}} & \multicolumn{3}{c|}{\textbf{$hasWatertight$}} & \multicolumn{3}{c}{\textbf{$hasWindResistance$}}& \multicolumn{3}{|c}{\textbf{$hasNumberOf\!Leaf$}} \\
\cline{2-16}
& P & R & F1 & P & R & F1 & P & R & F1 & P & R & F1& P & R & F1 \\
\hline
SVM & \textbf{1.00} & 0.29 & 0.44 & \textbf{1.00} & 0.55 & 0.71 & 0.86 & 0.50 & 0.63 & 0.83 & 0.45 & 0.59& \textbf{0.97} &	0.55&	0.70 \\
RF & \textbf{1.00} & \textbf{0.57} & \textbf{0.73} & \textbf{1.00} & 0.73 & 0.84 & \textbf{1.00} & 0.83 & 0.91 & \textbf{1.00} & 0.91 & 0.95& 0.93 &	0.87&	\textbf{0.90} \\
DT & 0.57 & \textbf{0.57} & 0.57 & 0.83 & \textbf{0.91} & \textbf{0.87} & \textbf{1.00} & \textbf{0.92} & \textbf{0.96} & 0.92 & \textbf{1.00} & \textbf{0.96}& 0.87&	\textbf{0.90}&	0.89 \\
KNN & \textbf{1.00} & 0.43 & 0.60 & 0.78 & 0.64 & 0.70 & 0.58 & 0.58 & 0.58 & 0.67 & 0.73 & 0.70 & 0.85&	0.67&	0.75\\
\hline
\end{tabular}
}
\end{table*}

\begin{table*}[h!]
\centering
\caption{Comprehensive performance of the RE models on the test corpus using syntactic and semantic features. The highest precision (P), recall (R) and F1-score (F1) are in bold.}\label{table10}
\resizebox{0.88\textwidth}{!}{
\begin{tabular}{lccc|ccc|ccc|ccc} 
\\
\multicolumn{13}{l}{Table 10. a: Comprehensive performance of the RE models on the test corpus using syntactic and semantic features.} \\
\hline
\multirow{3}{*}{\textbf{Models}} & \multicolumn{12}{c}{Relations} \\
\cline{2-13}
                                 & \multicolumn{3}{c}{$hasDimension$} & \multicolumn{3}{c}{$hasThermicCoe\!f\!ficient$} & \multicolumn{3}{c}{$hasAcousticAttenuation$} & \multicolumn{3}{c}{$hasFlameArrester$} \\ 
\cline{2-13}
& P & R & F1 & P & R & F1 & P & R & F1 & P & R & F1 \\
\hline
SVM & 0.90 & \textbf{0.78} & \textbf{0.84} & \textbf{1.00} & 0.75 & \textbf{0.86} & \textbf{1.00} & \textbf{0.76} & \textbf{0.86} & 0.67 & \textbf{0.67} & 0.67 \\
RF & \textbf{0.93} & 0.46 & 0.62 & \textbf{1.00} & 0.68 & 0.81 & \textbf{1.00} & 0.69 & 0.82 & \textbf{0.80} & \textbf{0.67} & \textbf{0.73} \\
DT & 0.72 & 0.73 & 0.73 & 0.81 & 0.75 & 0.78 & 0.85 & \textbf{0.76} & 0.80 & \textbf{0.80} & \textbf{0.67} & \textbf{0.73} \\
KNN & 0.85 & 0.76 & 0.80 & 0.96 & \textbf{0.79} & \textbf{0.86} & \textbf{1.00} & \textbf{0.76} & \textbf{0.86} & \textbf{0.80} & \textbf{0.67} & \textbf{0.73} \\
\hline
\end{tabular}}
\resizebox{\textwidth}{!}{
\begin{tabular}{lccc|ccc|ccc|ccc|ccc} 
\\
\multicolumn{16}{l}{Table 10. b: Comprehensive performance of the RE models on the test corpus using syntactic and semantic features.} \\
\hline
\multirow{3}{*}{\textbf{Models}} & \multicolumn{12}{c}{Relations}
\\ \cline{2-16}
& \multicolumn{3}{c|}{\textbf{$hasFireResistance$}} & \multicolumn{3}{c|}{\textbf{$hasAirPermeability$}} & \multicolumn{3}{c|}{\textbf{$hasWatertight$}} & \multicolumn{3}{c}{\textbf{$hasWindResistance$}}& \multicolumn{3}{|c}{\textbf{$hasNumberOf\!Leaf$}} \\
\cline{2-16}
& P & R & F1 & P & R & F1 & P & R & F1 & P & R & F1& P & R & F1 \\
\hline
SVM & 0.80 & 0.57 & 0.67 & 0.67 & 0.36 & 0.47 & \textbf{1.00} & 0.50 & 0.67 & 0.70 & 0.64 & 0.67& 0.86 &	0.80&	0.83 \\
RF & \textbf{1.00} & 0.43 & 0.60 & \textbf{1.00} & \textbf{0.55} & \textbf{0.71} & \textbf{1.00} & 0.67 & \textbf{0.80} & \textbf{1.00} & 0.91 & \textbf{0.95}& \textbf{0.96} &	0.78&	0.86 \\
DT & 0.67 & 0.57 & 0.62 & 0.86 & \textbf{0.55} & 0.67 & 0.69 & \textbf{0.75} & 0.72 & 0.85 & \textbf{1.00} & 0.92& 0.95&	\textbf{0.90}&	\textbf{0.92} \\
KNN & 0.86 & \textbf{0.86} & \textbf{0.86} & 0.75 & \textbf{0.55} & 0.63 & 0.88 & 0.58 & 0.70 & 0.90 & 0.82 & 0.86 & 0.91&	0.82&	0.86\\
\hline
\end{tabular}
}
\end{table*}

\begin{table}[ht!]
\centering
\caption{Comprehensive performance of relation extraction models using semantic and syntactic features, and using syntactic features only.}\label{table11}
\resizebox{0.7\textwidth}{!}{
\begin{tabular}{|l|p{1.5cm}|p{1.5cm}|p{1.5cm}||p{1.5cm}|p{1.5cm}|p{1.5cm}|}
\hline
\multirow{3}{*}{\textbf{Models}} & \multicolumn{6}{c|}{\textbf{Features}}\\
\cline{2-7}
& \multicolumn{3}{c||}{\textbf{Syntactic}} & \multicolumn{3}{c|}{\textbf{Semantic \& Syntactic}} \\
\cline{2-7} 
& \hspace{0.5cm}\textbf{P} & \hspace{0.5cm}\textbf{R} & \hspace{0.5cm}\textbf{F1} &\hspace{0.5cm} \textbf{P} & \hspace{0.5cm}\textbf{R} & \hspace{0.5cm}\textbf{F1} \\
\hline
SVM & \hspace{0.5cm}0.94 & \hspace{0.5cm}0.49 & \hspace{0.5cm}0.64 & \hspace{0.5cm}0.84 & \hspace{0.5cm}0.64 & \hspace{0.5cm}0.72 \\
RF & \hspace{0.5cm}\textbf{0.98} & \hspace{0.5cm}0.73 & \hspace{0.5cm}\textbf{0.83} & \hspace{0.5cm}\textbf{0.96} & \hspace{0.5cm}0.65 & \hspace{0.5cm}0.77 \\
DT & \hspace{0.5cm}0.80 & \hspace{0.5cm}\textbf{0.80} & \hspace{0.5cm}0.80 & \hspace{0.5cm}0.80 & \hspace{0.5cm}\textbf{0.74} & \hspace{0.5cm}0.76 \\
KNN & \hspace{0.5cm}0.82 & \hspace{0.5cm}0.62 & \hspace{0.5cm}0.70 & \hspace{0.5cm}0.87 & \hspace{0.5cm}0.73 & \hspace{0.5cm}\textbf{0.79} \\
\hline
\end{tabular}}
\end{table}

\subsubsection{Discussion}
Upon evaluating model performance with and without semantic features (Table \ref{table9} and Table \ref{table10}), clear differences were observed. DT and RF exhibited a slight decline in performance with semantic features across most relations, including $hasAirPermeability$. In contrast, KNN and SVM demonstrated improved precision, recall, and F1-scores for relations like "$hasThermicCoefficient$" and "$hasDimension$", while showing a negligible impact on "$hasFlameArrester$".

To determine the optimal RE model, Table \ref{table11} showcases the capacity of each classifier to extract all relations. RF emerged as the top performer, achieving an F1-score of 0.83 solely with syntactic features, underscoring its efficacy for this study and providing a clear direction for further optimization. Although DT exhibits better recall than RF, RF excels overall, prompting a focus on enhancing feature characteristics to improve recall, particularly addressing false negatives. Exploring ensemble methods are pivotal strategies to enhance performance further, especially given that results show each model performs well in some entities. Moreover, exploiting advanced word embeddings such as the Sentence-CamemBERT-Large Embedding Model for French may lead to the best results. This embedding model, along with "transformer embeddings", which not only embed individual words but also considers the context in which a word appears, represents the content and semantics of a French sentence in a mathematical vector. This enables a deeper understanding of the text beyond individual words in queries and documents, offering a powerful semantic search capability.

\section{Conclusion} \label{conclusion}
This research sought to enhance the efficacy of BIM in French construction projects through the automated extraction of technical requirements from French BTS documents. Given the importance of these documents for detailing essential technical specifications, the manual extraction process is not only time-consuming but also susceptible to errors. The methodological framework of the study encompassed several steps: 1) Gathering BTS documents from multiple online sources; 2) Implementing preprocessing strategies to eliminate irrelevant data such as headers and footers; 3) Segmenting BTS documents into raw requirements using their intrinsic numbering system; 4) Employing Transfer Learning techniques through fine-tuning of the “$Fr\_core\_news\_lg$” model and using the transformer-based “$Camem$ $BERT$”, which are particularly potent in the general French linguistic domain but previously untested in the construction-specific context. Furthermore, the research developed a rule-based system employing regular expressions and machine learning models, including CRF and BiLSTM-CRF, to benchmark against the advanced machine learning approaches. A novel validation method was introduced for the rule-based approaches, ensuring an equitable comparison between the traditional and modern machine learning methods.
For comprehensive requirement extraction, machine learning models were used to establish links and define relationships between entities. This utilized classifiers such as SVM, DT, RF, and KNN in conjunction with a detailed feature vector.
The findings indicated that Transfer Learning and transformer-based models substantially surpassed other methods in NER, achieving an F1-score exceeding 90\% across all entities. In RE, RF emerged as the most effective, with an F1-score exceeding 80\% for nearly all relationships. The outcomes will be used for further works to create a shape graph containing all extracted requirements, which aligns with a data graph extracted from the BIM model using SHACL (Shapes Constraint Language). This allows for the validation of the data graph against a set of constraints.

The study's contributions are multifaceted, significantly advancing the automation of requirement extraction and enhancing the reliability and efficiency of BIM implementations in the French construction sector. It not only underscored the utility of exploring French-specific models in construction but also expanded the scope of requirement extraction beyond mere entity identification to encompass full relationship and context understanding. By benchmarking against other approaches, the research affirmed the strengths of Transfer Learning and transformer-based techniques for domain-specific applications, setting a new benchmark for automation, accuracy, and comprehensiveness in requirement extraction within the industry.

\subsection{Limitations}
The study identified some key limitations. First, the proposed method was applied to non-scanned PDFs with a correct and consistent incremental numbering system. This methodology necessitates a preliminary analysis to verify the numbering system, which still involves manual effort. Additionally, while the segmentation system successfully extracted hierarchical structures from 72\% of the BTS documents, it encountered a considerable number of BTS with inconsistencies in numbering or a complete lack of a numbering system. These issues could potentially impact the segmentation algorithm, necessitating the development of adequate solutions to handle such cases. Another limitation is that the RE model relies on custom feature vector, which may affect the model's performance despite the approach demonstrating promising results. There is a continuous need to explore automated models that do not depend on the labor-intensive task of creating custom rules or features.

\subsection{Future Work}\label{future}
Although this study utilized LLMs, particularly MLMs like $CamemBERT$, future research will explore CLMs such as GPT\footnote{Generative Pre-trained Transformer (GPT) : \url{https://platform.openai.com/docs/models/gpt-4-turbo-and-gpt-4}} and Mistral\footnote{Mistral AI : \url{https://mistral.ai/fr/}}. These models have demonstrated profound understanding across various languages, including French. Utilizing prompt engineering may prove effective in NER and RE through zero-shot and few-shot learning approaches. This could potentially determine whether these techniques offer the best solution for domain-specific tasks without relying on large annotated datasets or manually crafted rules and features. For the segmentation algorithm, the plan is to leverage LLMs by providing instructions designed to extract the hierarchical structure of the BTS. This would utilize the intention mechanisms of the models, enabling a better understanding of context for such tasks. Additionally, the results obtained from this study will be filtered to remove redundancy in order to represent it as a knowledge graph, as previously mentioned, and to create a model that can automatically propose BIM products that are adequate for the extracted requirements.

\bibliographystyle{unsrt}
\bibliography{main_latex}
\end{document}